\newif\ifdraft
 \draftfalse

\documentclass[11pt]{article}

\usepackage[letterpaper,margin=1in]{geometry}
\usepackage[parfill]{parskip}

\usepackage{authblk}

\usepackage[toc,page,header]{appendix}

\usepackage[utf8]{inputenc} 
\usepackage[T1]{fontenc}    
\usepackage[colorlinks,citecolor=blue,urlcolor=blue,linkcolor=blue,linktocpage=true]{hyperref}       
\usepackage{url}            
\usepackage{booktabs}       
\usepackage{amsfonts}       
\usepackage{nicefrac}       
\usepackage{microtype}      
\usepackage{xcolor}         
\label{key}

\usepackage{comment}
\usepackage{amsmath}

\usepackage{epsfig}
\usepackage{graphicx}
\usepackage{wrapfig}
\usepackage{subfigure}
\usepackage{multirow}
\usepackage{hyperref}
\usepackage{graphicx}
\usepackage{caption}
\usepackage{array}
\usepackage{bbm}
\usepackage{color}
\usepackage{enumerate}
\usepackage{enumitem}
\usepackage{mathtools}
\usepackage{amsmath,amssymb,amsthm,bm}
\usepackage{thmtools}
\usepackage{amsfonts,graphicx}
\usepackage{mathrsfs}
\usepackage{amsmath,amssymb,amsfonts}
\usepackage{textcomp}
\usepackage{bbold}
\usepackage{graphicx}
\usepackage{textcomp}
\usepackage{xcolor}
\usepackage[ruled,vlined,linesnumbered]{algorithm2e}

\usepackage[authoryear]{natbib}
\usepackage{cleveref}

\usepackage[export]{adjustbox}

\newtheorem{remark-star}{Remark}
\newtheorem{remark-star-1}{Remark}

\newtheorem*{proof-sketch}{Proof Sketch}
\newtheorem{theorem}{Theorem}
\newtheorem{definition}{Definition}

\newcommand{\black}{\textcolor{black}}{}
\usepackage{booktabs}

\newcommand{\mem}{\mathrm{mem}}
\newcommand{\conf}{\mathrm{conf}}

\usepackage{amsthm}
\usepackage{amsmath}
\usepackage{amssymb}

\newcommand{\AlgName}{\textbf{ModelPred}\xspace}

\newcommand*{\ie}{\emph{i.e.},\@\xspace}

\newtheorem{innercustomthm}{Theorem}
\newenvironment{customthm}[1]{\renewcommand\theinnercustomthm{#1}\innercustomthm}
  {\endinnercustomthm}

\newcommand{\eps}{\varepsilon}
\newcommand{\R}{\mathbb{R}}
\newcommand{\g}{\nabla}
\DeclareMathOperator*{\argmin}{\arg\!\min}

\newcommand{\norm}[1]{\left\lVert#1\right\rVert}

\newcommand{\dparam}{d_{\mathrm{param}}}
\newcommand{\del}{\partial}
\newcommand{\A}{\mathcal{A}}

\newcommand{\bound}{B_1}
\newcommand{\boundthetastar}{B_2}

\newcommand{\Adnn}{\widehat{\A}}

\author[1]{Yingyan~Zeng}
\affil[1]{Virginia Tech}
\author[2]{Jiachen T. Wang}
\affil[2]{Princeton University}
\author[1]{Si~Chen}
\author[1]{Hoang Anh~Just}
\author[1]{Ran~Jin}
\author[1]{Ruoxi~Jia}

\title{ModelPred: A Framework for Predicting Trained Model from Training Data}
\begin{document}



\maketitle

\begin{abstract}
In this work, we propose \AlgName, a framework that helps to understand the impact of changes in training data on a trained model. This is critical for building trust in various stages of a machine learning pipeline: from cleaning poor-quality samples and tracking important ones to be collected during data preparation, to calibrating uncertainty of model prediction, to interpreting why certain behaviors of a model emerge during deployment. 
Specifically, \AlgName learns a parameterized function that takes a dataset $S$ as the input and predicts the model obtained by training on $S$.
Our work differs from the recent work of Datamodels \citep{ilyas2022datamodels} as we aim for predicting the trained model parameters directly instead of the trained model behaviors. 
We demonstrate that a neural network-based set function class is capable of learning the complex relationships between the training data and model parameters.
We introduce novel global and local regularization techniques to prevent overfitting and we rigorously characterize the expressive power of neural networks (NN) in approximating the end-to-end training process.
Through extensive empirical investigations, we show that \AlgName enables a variety of applications that boost the interpretability and accountability of machine learning (ML), such as data valuation, data selection, memorization quantification, and model calibration. 
\end{abstract}

\section{Introduction}\label{sec:intro}
\begin{center}
\noindent\fcolorbox{red}{yellow}{%
\begin{minipage}{0.9\columnwidth}
\emph{
What are the training points with the most or least contribution to the model? How to select data that benefit model performance? Is a training data point memorized by the model? How to produce accurate estimates of uncertainty on model predictions? Which point is the most responsible for learning a given parameter in the model? }
\end{minipage}
}
\end{center}

Answering these questions is central to building trust in machine learning (ML). 
Prior work shows that addressing these questions requires analyzing the input-output behavior of a learning process and gaining an understanding of how models (and the resulting predictions) might have differed if different training data points are observed~\citep{jia2019towards,ghorbani2019data,zhang2021counterfactual,efron1992bootstrap}.

However, due to the complexity of the learning process (e.g., end-to-end training), analyzing how its input---a dataset---affects the output---the trained model---is challenging. Recent work has made significant progress in approximating how a small change (e.g., removing one or a handful of training points) on the full training set would change the trained model~\citep{koh2017understanding,schioppa2022scaling,wu2020deltagrad,pruthi2020estimating}, but the existing techniques remain limited in estimating the effects of removing a large group of training points~\citep{koh2019accuracy}. \citet{ilyas2022datamodels} propose to learn a model that predicts the classification performance at a given test point directly from the training data. \black{However, this learning procedure will need to be redone if the classification performance at a different test point is of interest to the evaluation, which is common for practical applications with streaming-in testing data; therefore, it lacks the flexibility to accommodate different evaluation goals.}


In this paper, we propose a framework, \AlgName, to analyze the dependence of the trained models on training data by building an explicit model for the trained models in terms of training data. Compared to~\citep{ilyas2022datamodels}, \AlgName estimates the trained model parameters instead of the prediction performance at a certain test point, thereby providing flexibility to switch to different test points for evaluation. At the same time, the test performance estimates based on \AlgName are more accurate than those from \citep{ilyas2022datamodels} because \AlgName effectively leverages the knowledge of how test performance is calculated from a trained model.

Our \underline{contributions} are summarized as follows:

\noindent
\textbf{\#1: Introducing \AlgName.}
\AlgName is a framework designed to approximate the input-output behavior of a learning algorithm, $\A$, when applied to a training set $S$. The learning algorithm takes the training set and returns a trained model with parameters $\theta := \A(S)$. 
To accomplish this, \AlgName introduces a cheap-to-evaluate parametric model $\widehat{\A}$. We treat the search for the best $\widehat{\A}$ as a supervised learning problem. Specifically, we generate the labeled data for training $\widehat{\A}$ via executing $\A$ on different data subsets, and then use standard techniques in supervised learning to learn the best $\widehat{\A}$ that approximates $\A$. 
Our framework differs from the recent work of Datamodels \cite{ilyas2022datamodels} in the sense that we directly predict the trained model parameters instead of some behavior of the trained models. 

\noindent
\textbf{\#2: Instantiating \AlgName.} There are several critical design choices of \AlgName: What should be used as the input for $\widehat{\mathcal{A}}$? What is the right model class to learn $\widehat{\mathcal{A}}$? How to design a proper objective function to fit $\widehat{\mathcal{A}}$? 
\citet{ilyas2022datamodels} encodes the presence of each training sample of an input subset $S'$ within $S$ as a binary vector.
While this reduces the input complexity, this approach ignores the contents of an input subset. 
As a result, it can only estimate the outcome of learning for a subset \emph{within} the training dataset yet fails to estimate the learning result for, new \emph{unseen} points. 
In this paper, we use a neural-network-based \emph{set function} to approximate $\mathcal{A}$, which takes in the actual contents of a subset and predicts the trained model. This design enables the learned $\widehat{\mathcal{A}}$ to generalize to subsets involving unseen points. To mitigate overfitting, we propose two novel regularization techniques: the global regularizer encourages that the models predicted by the fitted $\widehat{\mathcal{A}}$ predict the labels in the held-out set as effectively as those obtained from the original $\mathcal{A}$; and the local one ensures that the gradient of the classification or regression learning objective function with respect to these models is close to zero.

\noindent
\textbf{\#3: Approximability of a learning algorithm with neural networks.} When instantiating \AlgName, we choose to approximate $\mathcal{A}$ with a neural network. While a neural network seems a reasonable first choice given its popularity in various vision and language tasks, it remains unclear whether a neural network can approximate complex training processes from end to end. This paper presents the \emph{first} formal study of the approximability of a learning process with neural networks. In particular, our results suggest that learning processes associated with strongly convex or smooth problems can be efficiently approximated by neural networks with ReLU units.

\noindent
\textbf{\#4: Applications of \AlgName.} We explore a variety of applications of \AlgName towards answering the questions posed at the beginning of the section. Our findings are summarized as follows.
\begin{itemize}
    \item \textbf{\AlgName functions successfully predict trained models from the training data (Table \ref{table:lrdel} and \ref{table:lradd}).} The Spearman correlation between the accuracy of the predicted models and the ground-truth is mostly greater than $97\%$.
    
     \item \textbf{\AlgName functions enable the discovery of the connection between data and model parameters (Figure \ref{fig:Sailency_MNIST}).} The level of sensitivity of a ground-truth model's parameter to different points matches the results obtained from a predicted model.
    \item \textbf{\AlgName functions successfully boost the accuracy of data valuation---measuring the contribution of individual points to learning (Table~\ref{table:shap3}).} While accurately measuring the value of data has a great potential to improve model performance and transparency, being more accurate comes with significant computational costs. In particular, existing data value notions require retraining a target model from scratch on many subsets of the training set and evaluating the corresponding model performance scores. We leverage the ability of \AlgName functions to predict models directly from training data and significantly improve the accuracy of data valuation.
    
    \item \textbf{\AlgName functions lead to improved data selection in the presence of bad data (Figure \ref{fig:SPAM_MNIST_Bad}).} We leverage the Shapley value (SV) estimated through \AlgName functions to rank the contributions of all training points and select the ones with the highest contributions. \AlgName functions lead to a significant improvement in data selection effectiveness while maintaining a similar computation cost to conventional Shapley value estimation methods.
    
    \item \textbf{\AlgName functions lead to more efficient and accurate quantification of model memorization (Figure \ref{fig:memory_acc}).} To assess whether a given training point is memorized or not, we evaluate \AlgName functions on different subsets with the point included and excluded and then compare the prediction at the point. We find that \AlgName functions enable the successful discovery of memorized training samples.
    
    \item \textbf{\AlgName functions improve the accuracy of prediction confidence (Table \ref{table:ece}).} We incorporate the predictions of the model's output by \AlgName functions into an ensemble, and it leads to an improvement in the accuracy of prediction confidence on various datasets.

\end{itemize}

\section{Approach}\label{sec:approach}
\newcommand{\x}{x}
\newcommand{\D}{\bm{D}}
\newcommand{\thetab}{\theta}

\subsection{Preliminaries}
Denote a data point as $(x, y)$, where $x\in \R^d$ is the feature vector and $y\in \R$ is the label. 
We define $f(x;\theta)$ as the \textbf{Base Model}, which is a parametric function that is parameterized by $\theta \in \R^{\dparam}$ and maps an input feature to a label (\ie logistic regression model).
The loss function $\ell(\theta; (x, y)) = \ell(f(x;\theta), y)$ is defined to measure the discrepancy between the prediction $\widehat y = f(x; \theta)$ and the true label $y$, which can be a classification or regression learning objective function.
Denote the full training dataset containing $n$ training data points as $D = \{(x_i, y_i)\}_{i=1}^n$.
We further denote $S$ as a subset with $n_S$ data points sampled from $D$. 
A \emph{learning algorithm} (\ie \emph{solver}) $\A$ is defined as the mapping from an input dataset $S$ to the parameters of the trained model $\widehat{\theta}_S$ \ie $\widehat \theta_S = \A(S)$. 
A typical example of a learning algorithm is empirical risk minimization (ERM) with $\ell_2$ penalty.
Specifically, the parameters are optimized to minimize the training loss averaged over the subset $S$:
\begin{equation}
\begin{aligned}
\widehat \theta_S := \argmin_{\theta } L(\theta;S)
= \frac{1}{n_S}\sum_{i \in S} \ell(\theta; (x_i, y_i)) + \frac{\lambda}{2} \norm{\theta}_2^2,
\end{aligned}
\label{equ:convex_prob}
\end{equation}
where $\lambda$ is a hyperparameter that controls the degree of penalization on the $\ell_2$-norm of the estimated parameters to increase the model's generalizability.
\subsection{Algorithm}
\subsubsection{Problem Formulation.}
As mentioned in Section~\ref{sec:intro}, our goal is to predict the outcome of the learning algorithm $\A$ given training set $S$.
While $\A(S)$ is a complicated function that typically involves loss function optimization, we leverage a classic technique in machine learning: picking a suitable function class and finding the function in the class that best approximates the target function. 
To this end, we construct \AlgName as a surrogate function $\Adnn$ to obtain outputs that match $\A(\cdot)$ given an arbitrary training set $S$. 

\newcommand{\Z}{\mathcal{Z}}

\begin{definition}[\AlgName function]
Given a data domain $\Z$,  a learning algorithm $\A$, a distribution $\mathcal{D}$ over the space of data \emph{sets} $2^\Z$, for any dataset $S$ drawn from $\mathcal{D}$, let $\widehat{\theta}_S = \A(S)$ be the model trained on $S$ using $\A$. 
The proposed \AlgName function is a parametric function $\Adnn$ optimized to predict the $\widehat{\theta}_S$ from a training set $S\sim \mathcal{D}$, \ie
\begin{align}
&\Adnn: 2^\Z \to \R^{\dparam}, \\ 
\mathrm{where~~~} 
&\Adnn = \argmin_{\widetilde{\A}} \mathbb{E}_{S \sim \mathcal{D}}
\left[L\left(\widetilde{\A}\left({S}\right), \A(S)\right)\right],
\end{align}
where $L(\cdot, \cdot)$ is a loss function.
\label{def:data2model}
\end{definition}

\subsubsection{Instantiating}
To develop a generic framework that  not only can predict a model from the subsets within the training dataset but also from \emph{unseen} subsets drawn from the same distribution $\mathcal{D}$, the contents of input data are preserved to be used as the input for \AlgName function, which makes $\A(\cdot)$ a set function. 
Taking this into consideration, we choose deep neural networks (DNNs) as the function class for \AlgName function $\Adnn$ given their strong expressive power. Moreover, since $\A(\cdot)$ is a set function, the architectures of neural networks are restricted to those that are invariant to input permutations. 
Finding the best neural network that approximates $\A(\cdot)$ becomes a \emph{supervised learning} problem when there are samples of $(S, \A(S))$ available. 

\subsubsection{Design of the Loss Function $L$}

Despite the strong expressiveness, with the large size of parameters, the DNN adopted as the surrogate model can be easily overfitted on the training set if we only push the model to minimize the discrepancy between the predicted and actual model parameters.
To prevent the DNN from overfitting, we propose two novel regularization techniques, where  the utility loss is proposed to maintain the utility of the predicted model as a global regularizer, while the optimality of the predicted parameters is imposed by the Karuch-Kuhn-Tucker (KKT) loss as a local regularizer.

\textbf{Global Regularizer.}
From an overall perspective, an accurately predicted model should maintain a utility close to the ground-truth model when evaluated on any test points, intuitively.
For the utility loss as a global regularizer, denote $\mathcal{U} (S;\widehat{\theta})$ as the utility function  that evaluates the utility of optimized model parameters $\widehat{\theta}$ on a dataset $S$.
For one training sample, $(\widehat{\theta}_S;S)$, the utility loss is specified as: 
$$L_{U}= \left| \mathcal{U} (S;\Adnn(S))-\mathcal{U} (S;\widehat{\theta}_S) \right|.$$

\textbf{Local Regularizer.}
\black{From a local perspective, an accurately predicted model should satisfy the optimality conditions, which constrains the gradient of the predicted parameters with respect to the learning objective function. With this in mind, we proposed to employ KKT conditions in the loss function to enhance the generalization of \AlgName by leveraging the optimality condition.}

\black{KKT conditions are found to guarantee the optimality of a solution \citep{bertsekas1997nonlinear} for convex problems.
For an unconstrained convex problem (\ref{equ:convex_prob}), the optimal parameters $\theta_S^* \in \argmin_{\theta \in \R^{\dparam}} L(\theta;S)$ should satisfy the stationary KKT condition as follows:
\begin{align}\label{eq:kkt}
\g L(\theta_S^*;S)=0.
\end{align}}
If the learning algorithm can find a near-optimal solution, then the stationary KKT condition is almost satisfied by the optimized parameter $\widehat \theta_S$.
Therefore, the KKT loss of one training sample is added as: $L_{KKT}= \norm{ \nabla L\left(\Adnn(S);S\right) }$.
This condition also applies to convex problems with hard constraints where the KKTstationary condition can be enforced on the Lagrangian of the optimization problem.
Besides, the stationary KKT condition can be treated as the first-order optimality condition for nonconvex optimization problems.
Therefore, by adding KKT loss, we enhance the generalization of \AlgName for generic optimization problems, including neural networks.

Combining the loss on the discrepancy between prediction and groud-truth with proposed regularizers, the total loss of one training sample for the proposed DNN is:
\begin{equation}
\begin{aligned}
    L_{\mathrm{DNN}}&=\norm{\widehat{\theta}_S -\Adnn(S)} + L_{KKT} + L_U.
\end{aligned}
\end{equation}

\subsubsection{Algorithm Details}
\begin{figure*}[ht]
\includegraphics[width=\textwidth]{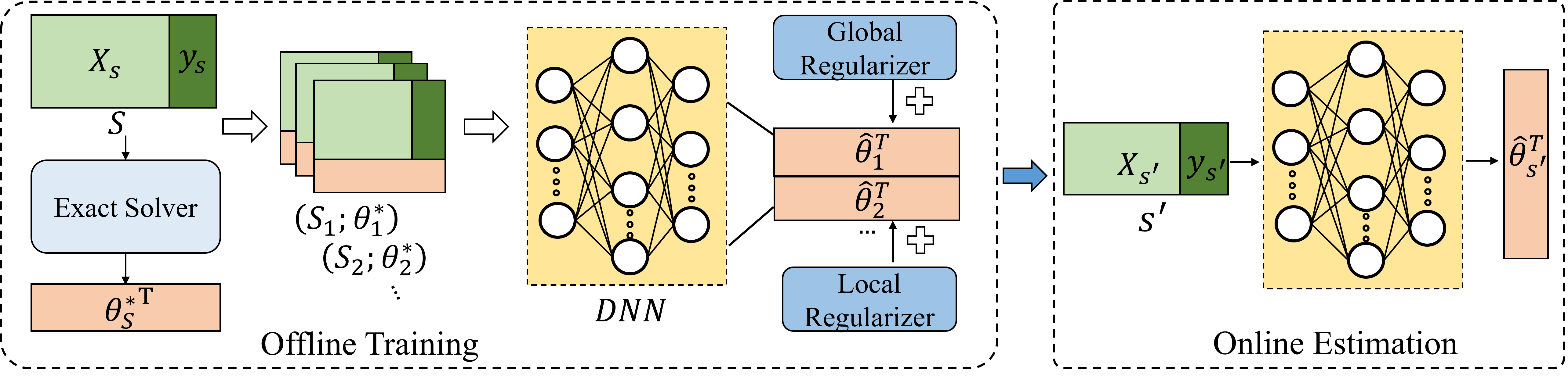}
\caption{The offline training phase and online estimate of the proposed \AlgName.}
\label{fig:overview}
\end{figure*}

The proposed \AlgName framework proceeds in two phases: \textit{offline training} and \textit{online estimation} (Figure~\ref{fig:overview}.)


\textbf{Offline Training.}
The offline training of \AlgName consists of the parameters sampling step and DNN training step.
With the set of training samples $\Phi = \{(S_1,\widehat{\theta}_1),(S_2,\widehat{\theta}_2),\dots \}$, the objective of the offline training phase is to train an effective parameter model which can predict the optimized parameters accurately given new input data points.

\emph{
The very first question encountered during the training phase is how to generate training subsets and what distribution should be used.}

Denote $\pi$ as a permutation of all training data points from the full set and $D_{\pi[:i]}$ as a set of points that precedes the $i$-th point in $\pi$.
We adopt permutation sampling to sample the trained parameters $\A(D_{\pi[:i]})$ for training DNN-based $\Adnn$. 
The offline training workflow is summarized in Algorithm~\ref{alg:permutationsampling}.

\begin{algorithm}[ht]
\fontsize{9}{9}\selectfont
\SetAlgoLined
\SetKwInOut{Input}{input}
\SetKwInOut{Output}{output}
\Input{Full dataset $D = \{(x_i, y_i)\}_{i=1}^n$, learning algorithm $\A$, the number of permutations $T$}
\Output{Trained $\Adnn$}

$\Phi$ $\xleftarrow{}$ $\emptyset$

\For{$t = 1, \dots, T$}{
   $\pi_t$  $\xleftarrow{}$ GenerateUniformRandomPermutation($D$)\\
   \For{$i=1, \dots n$}{
   ${\widehat {\theta}_i^{\pi_t}}= \A(D_{{\pi_t}[:i]})$\\ 
   $\Phi = \Phi\cup\{(D_{{\pi_t}[:i]};\widehat {\theta}_i^{\pi_t})\}$
   }
}
Trained $\Adnn$ with $\Phi$

\Return{$\Adnn$}
\caption{\AlgName Offline Training}
\label{alg:permutationsampling}
\end{algorithm}

Compared with other kinds of subset sampling strategies such as sampling uniformly at random, permutation sampling enables the evaluation of the change of model parameters with slightly varied subsets.
Therefore, the training samples collected by permutation can better reflect the effect of individual data points, thus, allowing the DNN to better learn their influences in the trained parameters.
Thus, permutation sampling is used to generate the training subsets in the offline training of \AlgName.
The empirical experiment results show a slight advantage of permutation sampling over uniform sampling when we evaluate the performance of the DNN trained by subsets.
Note that if the distribution of the subset to be estimated in the testing phase (\ie test distribution) is known a prior, it is suggested to directly sample the training samples from the test distribution.
For instance, in data valuation by SV, the definition of SV requires the distribution to assign uniform probability to all data sizes and uniform probability to all subsets of a given size. Then, in the offline phase, the subsets should be sampled from the distribution above.



%


\textbf{Online Estimation.}
After the offline training phase, the trained DNN $\Adnn$ can be used for efficient model parameters prediction given a new training subset or a batch of subsets through an evaluation of $\Adnn$.
Note the subsets are not limited to those containing observed data points.

\section{Characterization of Efficiently Approximatable Learning Algorithms}\label{sec:theory}

In this paper, we try to use a neural network to fit the function $\widehat \theta = \A(D)$. 
When the context is clear, we omit the learning algorithm and simply write the function as $\widehat \theta(D)$. We denote the function associated with $k$th parameter as $\widehat \theta_k(D)$, and thus $\widehat \theta(D) = (\widehat \theta_1(D), \ldots, \widehat \theta_{\dparam}(D))$, where $\dparam$ is the number of model parameters. Although $D$ is a set of data points, it can be equivalently viewed as a vector concatenation of all $(x_i, y_i)$. We assume we always normalize input features to $[0, 1]$. Therefore, we have $D \in [0, 1]^{n(d+1)}$. 
In order to fit $\widehat \theta(D)$ with a neural network, the very first question is \emph{whether $\widehat \theta(D)$ could be efficiently expressed or approximated by neural networks of certain structures. }
Here, the efficiency is measured by the total number of computational units in the neural networks.

Despite the strong expressive power of neural networks, not every function could be efficiently approximated by them. Particularly, while the famous universal approximation theorem (UAT) includes many functions, there are requirements regarding the continuity of the functions and compactness of the domain. A recent study \citep{yarotsky2017error} shows that \emph{functions $g: \R^d \rightarrow \R$ that have a smaller upper bound of gradient norm $\norm{\g g}$ could be more efficiently approximated by neural networks with ReLU activation}. As long as $\widehat \theta_k(D)$ could be efficiently approximated by certain neural network architectures, we could put $\dparam$ such networks in parallel to approximate $\widehat \theta(D)$. The parallel networks share the same input but have no internal connections or computational units. Therefore, we can \emph{reduce the problem of understanding the efficient approximability of a function to that one of bounding its gradient norm}. Particularly, this section aims to understand \emph{sufficient} conditions for which $\left\| \frac{\del \widehat \theta_k}{\del D} \right\|$ could be upper bounded. 
 
Our first result is under the setting that the learning algorithm $\A$ is able to find an \emph{optimal parameter} $\theta^* \in \argmin_\theta L(\theta; D)$. There are three major assumptions in our result: (1) the loss function $\ell(\theta)$ is $\alpha$-strongly convex, (2) $\norm{\frac{\del}{\del \theta \del z_i} \ell (\theta, z)}$ is upper bounded by some constant $\bound$, and (3) $\norm{\theta^*} \le \boundthetastar$. 

\begin{definition}[$\alpha$-strongly convex]
A differentiable function $f$ is $\alpha$-strongly convex if 
\begin{align}
    f(y) \ge f(x) + \g f(x)^T (y-x) + \frac{\alpha}{2} \norm{y-x}^2,
\end{align}
for some $\alpha > 0$ and for all $x, y$ in the domain. 
\label{def:strongly-convex}
\end{definition}

Since $\ell(\theta; (x, y))$ is $\alpha$-strongly convex, the training loss $L(\theta;D) = \frac{1}{n} \sum_{i=1}^n \ell(\theta; (x_i, y_i)) + \frac{\lambda}{2} \norm{\theta}_2^2$ is $(\alpha+\lambda)$-strongly convex. Together with the condition that $\norm{\theta^*}$ is finite, it implies $L$ has a unique global minimum $\theta^* = \argmin_\theta L(\theta; D)$. Our main result for strongly convex functions is below:
\begin{theorem}
If for all $z \in [0, 1]^{d}$, the loss function $\ell(\theta; z)$ is $\alpha$-strongly convex in $\theta$, and $\norm{\frac{\del}{\del \theta \del z_i} \ell (\theta, z)} \le \bound$ for $i \in [d]$, 
then for $\theta_k (D) = [\argmin_\theta L(\theta; D)]_k$, where $L(\theta; D) = \frac{1}{n} \sum_{i=1}^n \ell(\theta; (x_i, y_i)) + \frac{\lambda}{2} \norm{\theta}_2^2$ and $[\cdot]_k$ means the $k$th element in the vector, then we have 
\begin{align}
    \norm{\frac{\del \theta_k}{\del D}}
    \le \bound \frac{\sqrt{\dparam (d+1)}}{\sqrt{n} (\alpha + \lambda)}.
\end{align}
\label{thm:stronglyconvex}
\end{theorem}

We then analyze the case where the learning algorithm $\A$ is gradient descent. Gradient descent is a commonly used optimization algorithm in machine learning. Formally, suppose the parameter $\theta$ is initialized by $0$ and denoted by $\theta^{(0)}$. For the $i$th iteration, we update $\theta^{(t)} = \theta^{(t-1)} - \eta \g L(\theta^{(t-1)}; D)$ with a learning rate $\eta$. Given the maximum step number $T$, we have $\widehat \theta(D) = \theta^{(T)}$. 
The only different assumption for bounding $\norm{\frac{\del \widehat \theta_k}{\del D}}$ in this case is that the loss function $\ell(\theta)$ is $\beta$-smooth instead of $\alpha$-strongly convex. 
\begin{definition}[$\beta$-smoothness]
A differentiable function $f$ is $\beta$-smooth if 
\begin{align}
    \norm{\g f(x) - \g f(y)} \le \beta \norm{x-y}
\end{align}
for some $\beta > 0$, and for all $x, y$ in the domain. 
\label{def:smoothness}
\end{definition}
\begin{theorem}
If for all $z \in [0, 1]^{d}$, the loss function $\ell(\theta; z)$ is $\beta$-smooth in $\theta$, and $\norm{\frac{\del}{\del \theta \del z_i} \ell (\theta, z)} \le \bound$ for $i \in [d]$, 
then for $ \theta_k^{(t)}(D)$ defined by the $k$th entry of $ \theta^{(t)}$, which is iteratively computed by $\theta^{(t)} = \theta^{(t-1)} - \eta \g L(\theta^{(t-1)}; D)$ and $\theta^{(0)}=0$ with a learning rate $\eta > 0$, and $L(\theta; D) = \frac{1}{n} \sum_{i=1}^n \ell(\theta; (x_i, y_i)) + \frac{\lambda}{2} \norm{\theta}_2^2$, we have 
$$
\norm{\frac{\del \theta_k^{(t)}}{\del D}} 
\le \left( 1-(1-\eta \lambda -\eta \dparam \beta)^t \right) \frac{B_1 \sqrt{(d+1)} }{ \sqrt{n}(\lambda+\dparam\beta) }.
$$
\label{thm:smooth}
\end{theorem}

Theorem \ref{thm:stronglyconvex} and \ref{thm:smooth} demonstrate that $\left\| \frac{\del \widehat \theta_k}{\del D} \right\|$ could be upper bounded the dimension of data domain $d$ and model parameter dimension $\dparam$, and the dependency is only $O(\sqrt{\dparam d / n})$ for the case of $\A$ being able to find the optimal parameter, and $O(\sqrt{d}/\dparam \sqrt{n})$ for the case of gradient descent.




\section{Experiments}\label{sec:exp}
\subsection{Experimental setup} 
\subsubsection{Computational Considerations} 
The retraining of a large base model, such as a neural network on a large dataset is computationally intensive.
To apply the proposed method efficiently to large base models, we propose to employ the technique of transfer learning, where the weights in all layers except the last one are fixed as a feature extractor \citep{sun2019meta}.
To retrain a neural network on large size subsets, only the last layer in the neural network will be modified and adjusted to the training dataset, 
This technique has been widely used in natural language processing and computer vision domains~\citep{yuan2021florence}.
Besides, by removing the data loading bottleneck, \emph{fast neural network training} techniques such as FFCV\citep{leclerc2022ffcv} make it possible to train large amouts of models on diverse subsets in an efficient manner.
Our paper does not incorporate such advanced techniques, and we expect that doing so can further improve efficiency. 

\subsubsection{Base Model} 
To demonstrate that our proposed \AlgName is a model-agnostic approach, three kinds of base models (\ie Logistic Regression (LR), Support Vector Machine (SVM)), and a neural network (Resnet-18)) are chosen, and \AlgName will learn to predict the parameters of these models.
For LR and SVM, a regularization term with $\ell_2$-norm coefficient $\lambda=1$ is added to the loss function. 
We adopt L-BFGS-B \citep{zhu1997algorithm} as the learning algorithm (\ie solver) for LR and LIBLINEAR \citep{fan2008liblinear} as the solver for SVM with a squared hinge loss.
Stochastic Gradient Descent (\ie SGD with learning rate as 0.001, and momentum as 0.9) is adopted as the solver  for the training of NN.

\subsubsection{Dataset} We evaluate the proposed \AlgName on five datasets: 

\textbf{Iris (\citep{misc_iris_53}).} We use the binary version of Iris dataset with the first two classes of data. 
The binary version contains 100 samples in total with feature dimension $d=4$. 
67 data points are randomly selected to generate training subsets.
The remaining 33 data points are used for SV calculation, and 17 of them are used for dataset addition.

\textbf{SPAM (\citep{misc_spambase_94}).} SPAM dataset  is a collection of spam and non-spam e-mails with feature dimension $d=215$.
300 data points are randomly selected to generate training subsets.
500 data points are used as the testing set for SV calculation, and 150 of them are used for dataset addition.

\textbf{HIGGS (\citep{baldi2014searching}).} HIGGS dataset is  produced using Monte Carlo simulations with feature dimension $d=30$.
After preprocessing, we keep 25 features.
300 data points are randomly selected to generate training subsets.
500 data points are used as the testing set for SV calculation, and 150 of them are used for dataset addition.

\textbf{MNIST (\citep{lecun1998gradient}).} MNIST dataset is a collection of grayscale handwritten digits with size $28\times 28$ and 10 classes.
We use a CNN in the preprocessing stage to extract the features and reduce the dimension to 128.
\black{300 data points are randomly selected to generate training subsets.
500 data points are used as the testing set for SV calculation, and 150 of them are used for dataset addition.}

\textbf{CIFAR-10 (\citep{krizhevsky2009learning}) and Hymenoptera (\citep{paszke2019pytorch}).} CIFAR-10 is a collection of 60,000 3-channel images in 10 classes. 
Hymenoptera is a small 3-channel  dataset  used to classify ants and bees, which consists of 245 training images and 153 testing images.
These two datasets are used for applying \AlgName to large NNs by transfer learning.
In the experiment, we fine-tune the weights of the last layer of a pre-trained Resnet-18 model on CIFAR-10 dataset for the binary classification of hymenoptera images and use \AlgName to estimate the parameters of the new model. 
In particular, a Resnet-18 model is pre-trained on CIFAR-10.
The dimension of the weight in the last layer is 512 (\ie $d=512$).
208 images are randomly selected to generate training subsets and the remaining 190 are treated as the testing set, where 100 of them are used for dataset addition.

All the input data $x$ are normalized to rescale the norm $\norm{x}$ to be within the range of $[0, 1]$.

\begin{table*}
\centering
\resizebox{\textwidth}{!}{%
\begin{tabular}{c|cccccccc}
\toprule
\multirow{2}{*}{Dataset} & \multicolumn{1}{c}{\multirow{2}{*}{\shortstack{Total Training \\ Data Points}}} & \multirow{2}{*}{\shortstack{Total Testing \\ Data Points}} & \multicolumn{3}{c}{Dataset Deletion} & \multicolumn{3}{c}{Dataset Addition} \\ \cmidrule{4-9} 
 & \multicolumn{1}{c}{} &  & \shortstack{Size of \\Starting Subset} & \shortstack{Size of \\ Deleted Subsets} & \shortstack{Size of \\ Ending Subset} & \shortstack{Size of \\ Starting Subset} & \shortstack{Size of \\Added Subsets} & \shortstack{Size of \\ Ending Subset} \\ \midrule
Iris & 67 & 33 & 63 & {[}5, 10, 15, …, 30{]} & 33 & 67 & {[}1, 2, 3, …, 17{]} & 84\\
\black{SPAM, HIGGS, MNIST} & 300 & 500 & 300 & {[}5, 10, 15, …, 150{]} & 150 & 150 & {[}5, 10, 15, …, 150{]} & 300\\ 
Hymenoptera & 208 & 190 & 208 & {[}5, 10, 15, …, 100{]} & 108 & 108 & {[}5, 10, 15, …, 100{]} & 208 \\ \bottomrule
\end{tabular}
}
\caption{Setting for Dataset Deletion and Dataset Addition.}
\label{table:deladdsetting}
\end{table*}

\subsubsection{Sampling Distribution}
15000 subsets are sampled from each dataset following permutation sampling procedures in Algorithm~\ref{alg:permutationsampling} to construct the training sample set $\Phi$.

\subsubsection{Proposed Network Structure}
\black{We adopt a canonical model architecture DeepSets \citep{zaheer2017deep} for proposed DNN. 
DeepSets is designed to be permutation invariant of the input samples, which is fit for set function learning (\ie mapping a set of samples to a target output).
A DeepSet model is a \emph{set} function $f(S)=\rho(\sum_{x\in S}\phi(x))$ where both $\rho$ and $\phi$ are neural networks.
In the experiment, both $\rho$ and $\phi$ networks have 3 fully-connected layers with 128 neurons in each layer.}

\subsubsection{Evaluation Metrics and Baseline}
To evaluate the performance of the proposed \AlgName, baseline comparisons are conducted in two main scenarios: \emph{dataset deletion} and \emph{dataset addition}, where \AlgName is used to predict the model on deleted or added datasets.
During the deletion, the size of training subsets is decreased from 100\% of the full size to 50\%  (i.e, for SPAM, from 300 to 150).
During the addition, the size of training subsets is increased from 50\% of the full size to 100\% (i.e, for SPAM, from 150 to 300).
Notably, the added samples are unseen during the offline training phase.
In both scenarios, 10 subsets are generated randomly with each size.
To illustrate the generalizability of the proposed \AlgName to new data points from the same distribution, previously unseen data points are incorporated to form a larger subset in the dataset addition scenario. 
Table~\ref{table:deladdsetting} details the size of the added or deleted subsets for each dataset for dataset deletion and dataset addition.

In these two scenarios, \textbf{Influence Function} is selected as one baseline.
With the ability to approximate the effect of single data point deletion or addition on model parameters, Influence Function can also be extended to evaluate the subset change \citep{koh2019accuracy}. 
\black{\textbf{Datamodel} \citep{ilyas2022datamodels} is selected as another baseline only for dataset deletion since it cannot make predictions for newly added training points.}
Besides, DeltaGrad \citep{wu2020deltagrad} can also serve as a baseline for batch deletion or addition. However, it requires the application of SGD to optimize the base model, which cannot provide the optimal solution in a timely manner and cannot solve the primal problem of SVM accurately. 
Therefore, we do not include DeltaGrad in the experiments.
Additionally, we conduct the ablation study by creating \textbf{\textbf{ParaLearn}\xspace} as another baseline.
\textbf{ParaLearn} has the same neural network structure as \AlgName and also predicts the optimized parameter for a convex learning model. 
The difference is that it does not include local and global regularizers.

Three metrics are adopted to evaluate the effectiveness of the proposed method: 1) \textbf{the Euclidean distance} between the estimated model parameters and the exact parameters $\theta^*$ provided by the solver, \ie $\norm{\widehat{\theta}-\theta^*}$;
2) \textbf{the Normalized-Root-Mean-Squared Error (NRMSE)} of the estimated utility calculated by the estimated model parameters;
3) \textbf{the Spearman rank-order correlation} (\citep{spearman}) between the ground-truth utility and the estimated utility on the testing set.
The first metric evaluates the accuracy of parameter estimation.
Here, we use the parameters optimized by retraining the model with the solver as the ground-truth.
We adopt the Spearman correlation in addition to the NRMSE because most of the applications of subsets utility \citep{akhbardeh2019multi} desire that the utility is ranked in the correct order.



\subsubsection{Machine configuration}
We run experiments with one Intel(R) Xeon(R) Gold 5218 CPU and use one
GeForce RTX 2080 Ti for DNN training and inference.

\subsection{Experimental results}\label{subsec:exp}

\subsubsection{\AlgName can predict training outcomes}
\begin{table*}
\centering
\scalebox{0.75}{
\begin{tabular}{c|c|cc|cc|cc}
\toprule
\multirow{2}{*}{Dataset} &
  \multirow{2}{*}{Algorithm} &
  \multicolumn{2}{c|}{Parameter} &
  \multicolumn{4}{c}{Utility}\\ \cmidrule{3-8} 
 &
   &
  $\norm{\widehat{\theta}-\theta^*}$ &
  Std &
  NRMSE &
  Std &
  Spearman Corr &
  Std \\ \midrule
\multirow{4}{*}{Iris} &
  \textbf{\AlgName} &
  \textbf{9.67E-02} &
  1.51E-02 &
  \textbf{2.80\%} &
  \textbf{0.42\%} &
  \textbf{0.9964} &
  \textbf{0.0107} \\
 &
  Influence function &
  8.97E-01 &
  \textbf{1.39E-02} &
  74.62\% &
  16.98\% &
  0.4071 &
  0.4262 \\
 &
  ParaLearn &
  1.95E-01 &
  2.49E-02 &
  9.47\% &
  2.61\% &
  0.9679 &
  0.0297 \\
  
 &
  \black{Datamodel} &
  N/A &
  N/A &
  22.78\% &
  5.72\% &
  0.9571 &
  0.0474 \\
 \midrule
\multirow{4}{*}{SPAM} &
  \textbf{\AlgName} &
  \textbf{3.72E-01} &
  1.27E-02 &
  \textbf{6.48\%} &
  \textbf{1.19\%} &
  \textbf{0.9856} &
  \textbf{0.0040} \\
 &
  Influence function &
  1.07E+00 &
  \textbf{3.40E-03} &
  50.60\% &
  3.70\% &
  0.1841 &
  0.1797 \\
 &
  ParaLearn &
  2.36E+00 &
  2.15E-02 &
  125.55\% &
  10.62\% &
  N/A &
  N/A \\
 &
  \black{Datamodel} &
  N/A &
  N/A &
  69.53\% &
  9.98\% &
  0.6174 &
  0.0612 \\
  \midrule
\multirow{4}{*}{HIGGS} &
  \textbf{\AlgName} &
  \textbf{2.41E-01} &
  1.77E-02 &
  \textbf{12.58\%} &
  \textbf{3.86\%} &
  \textbf{0.7980} &
  \textbf{0.0662} \\
 &
  Influence function &
  3.32E-01 &
  \textbf{5.11E-03} &
  22.68\% &
  6.11\% &
  0.7504 &
  0.1275 \\
 &
  ParaLearn &
  5.11E-01 &
  2.89E-02 &
  65.30\% &
  17.03\% &
  N/A &
  N/A \\ 

 &
  \black{Datamodel} &
  N/A &
  N/A &
  167.83\% &
  62.34\% &
  0.3414 &
  0.1412 \\
\midrule
\multirow{4}{*}{MNIST} &
  \textbf{\AlgName} &
  \textbf{9.92E-01} &
  9.60E-03 &
  \textbf{2.28\%} &
  \textbf{0.42\%} &
  \textbf{0.9978} &
  \textbf{0.0010} \\
 &
  Influence function &
  1.04E+01 &
  6.65E-01 &
  48.85\% &
  2.81\% &
  0.0278 &
  0.2375 \\
 &
  ParaLearn &
  4.58E+00 &
  \textbf{4.96E-03} &
  113.94\% &
  7.11\% &
  N/A &
  N/A  \\
 &
  \black{Datamodel} &
  N/A &
  N/A &
  46.30\% &
  3.12\% &
  0.9866&
  0.0058 \\
\midrule
\multirow{4}{*}{Hymenoptera} &
  \textbf{\AlgName} &
  \textbf{1.18E+00} &
 1.14E-02 &
  \textbf{6.54\%} &
  \textbf{0.53\%} &
  \textbf{0.9920} &
  \textbf{0.0025} \\
 &
  Influence function &
  1.24E+00 &
  \textbf{6.24E-03} &
  56.43\% &
  2.71\% &
  0.03623 &
  0.13195 \\
 &
  ParaLearn &
  2.04E+00 &
  1.16E-02 &
  73.97\% &
  4.05\% &
  N/A &
  N/A \\
  &
  \black{Datamodel} &
  N/A &
  N/A &
  51.83\% &
  4.86\% &
  0.8042&
  0.0622 \\
 
 \bottomrule
\end{tabular}
}
\caption{A summary of \AlgName results and baseline comparison results in the scenario of dataset deletion with LR (\ie for Iris, SPAM, HIGGS, and MNIST) and NN (\ie for Hymenoptera) as the base model. Mean and standard deviation reported over 10 experimental trials. The best result is highlighted in \textbf{bold}.}
\label{table:lrdel}
\end{table*}

\begin{table*}
\centering
\scalebox{0.75}{
\begin{tabular}{c|c|cc|cc|cc}
\toprule
\multirow{2}{*}{Dataset} &
  \multirow{2}{*}{Algorithm} &
  \multicolumn{2}{c|}{Parameter} &
  \multicolumn{4}{c}{Utility}\\ \cmidrule{3-8} 
 &
   &
  $\norm{\widehat{\theta}-\theta^*}$ &
  Std &
  NRMSE &
  Std &
  Spearman Corr &
  Std \\ \midrule
\multirow{4}{*}{Iris} &
  \textbf{\AlgName} &
  \textbf{1.34E-01} &
  \textbf{8.81E-03} &
  \textbf{5.10\%} &
  \textbf{0.52\%} &
  \textbf{0.9882} &
  \textbf{0.0067} \\
 &
  Influence function &
  3.56E-01 &
  6.13E-04 &
  63.07\% &
  3.15\% &
  0.0779 &
  0.1521 \\
 &
  ParaLearn &
  2.85E-01 &
  4.22E-03 &
  43.91\% &
  2.83\% &
  0.9730 &
  0.0161 \\ 
 \midrule
\multirow{4}{*}{SPAM} &
  \textbf{\AlgName} &
  \textbf{5.96E-01} &
  \textbf{4.56E-03} &
  \textbf{10.22\%} &
  \textbf{0.82\%} &
  \textbf{0.9919} &
  \textbf{0.0026}\\
 &
  Influence function &
  1.11E+00 &
  2.34E-03 &
  65.44\% &
  2.26\% &
  0.1886 &
  0.1683 \\
 &
  ParaLearn &
  2.19E+00 &
  2.10E-02 &
  145.73\% &
  4.36\% &
  N/A &
  N/A\\ \midrule
\multirow{4}{*}{HIGGS} &
  \textbf{\AlgName} &
  \textbf{3.26E-01} &
  \textbf{1.56E-02} &
  \textbf{14.11\%} &
  \textbf{2.71\%} &
  \textbf{0.8054} &
  \textbf{0.0684} \\
 &
  Influence function &
  4.58E-01 &
  3.23E-03 &
  20.40\% &
  4.27\% &
  0.9008 &
  0.0381 \\
 &
  ParaLearn &
  1.92E+00 &
  1.86E-02 &
  838.62\% &
  128.59\% &
  N/A &
  N/A \\ \midrule
\multirow{4}{*}{MNIST} &
  \textbf{\AlgName} &
  \textbf{1.81E+00} &
  \textbf{1.22E-02} &
  \textbf{5.46\%} &
  \textbf{0.25\%} &
  \textbf{0.9969} &
  \textbf{0.0009} \\
 &
  Influence function &
  1.31E+01 &
  4.75E-01 &
  125.26\% &
  4.22\% &
  -0.8431 &
  0.0367 \\
 &
  ParaLearn &
  4.76E+00 &
  6.01E-03 &
  131.73\% &
  2.17\% &
  N/A &
  N/A \\ \midrule
\multirow{4}{*}{Hymenoptera}  &  
\textbf{\AlgName} &
  \textbf{1.94E+0} &
  1.57E-02 &
  \textbf{17.45\%} &
  0.58\% &
  \textbf{0.70000} &
  \textbf{0.09458} \\
 &
  Influence function &
  7.11E+00 &
  2.81E-02 &
  558.29\% &
  8.77\% &
  0.35000&
  0.14044 \\
 &
  ParaLearn &
  2.25E+00 &
  \textbf{1.37E-02} &
  27.13\% &
  \textbf{0.24}\% &
  N/A &
  N/A\\ \bottomrule
\end{tabular}
}
\caption{A summary of \AlgName results and baseline comparison results in the scenario of dataset addition with LR as the base model. Mean and standard deviation reported over 10 experimental trials. The best results are highlighted in \textbf{bold}.}
\label{table:lradd}
\end{table*}

Table \ref{table:lrdel} and \ref{table:lradd} summarize results of proposed \AlgName and the baseline in scenarios of dataset deletion and addition, respectively.
LR is used as the base model for the first four datasets while NN is used for the last Hymenoptera dataset.
First, we focus on the results with LR as the base model.
In general,  \AlgName demonstrates a significant advantage in the accurate prediction of parameters if we compare $\norm{\widehat{\theta}-\theta^*}$.
\textbf{Comparing the results of different datasets, the Euclidean distance grows slightly with the dimension of the input feature dimension, but the estimated utility is unaffected.} 
It can also be shown the predicted parameters well maintain the utility by checking the NRMSE of utility and the Spearman correlation (\ie over 95\% for most datasets) between the estimated utility and ground-truth.
However, it is seen that both the parameter prediction accuracy and the utility estimation accuracy are relatively low on the HIGGS dataset.
This might be caused by the low performance of the base model (\ie LR) on this dataset, which generates similar optimal parameters as training samples in the training phase.

By investigating the performance of three baselines, although \textbf{ParaLearn}  outperforms the Influence Function on Iris dataset,
the Spearman correlation for other datasets cannot be calculated.
This is because it provides constant prediction regardless of the input subset change, resulting in constant utility for all subsets. 
\textbf{This result indicates the proposed KKT and utility loss effectively prevent overfitting.}



\begin{figure}[ht]
\centering
\includegraphics[width=0.9\columnwidth]{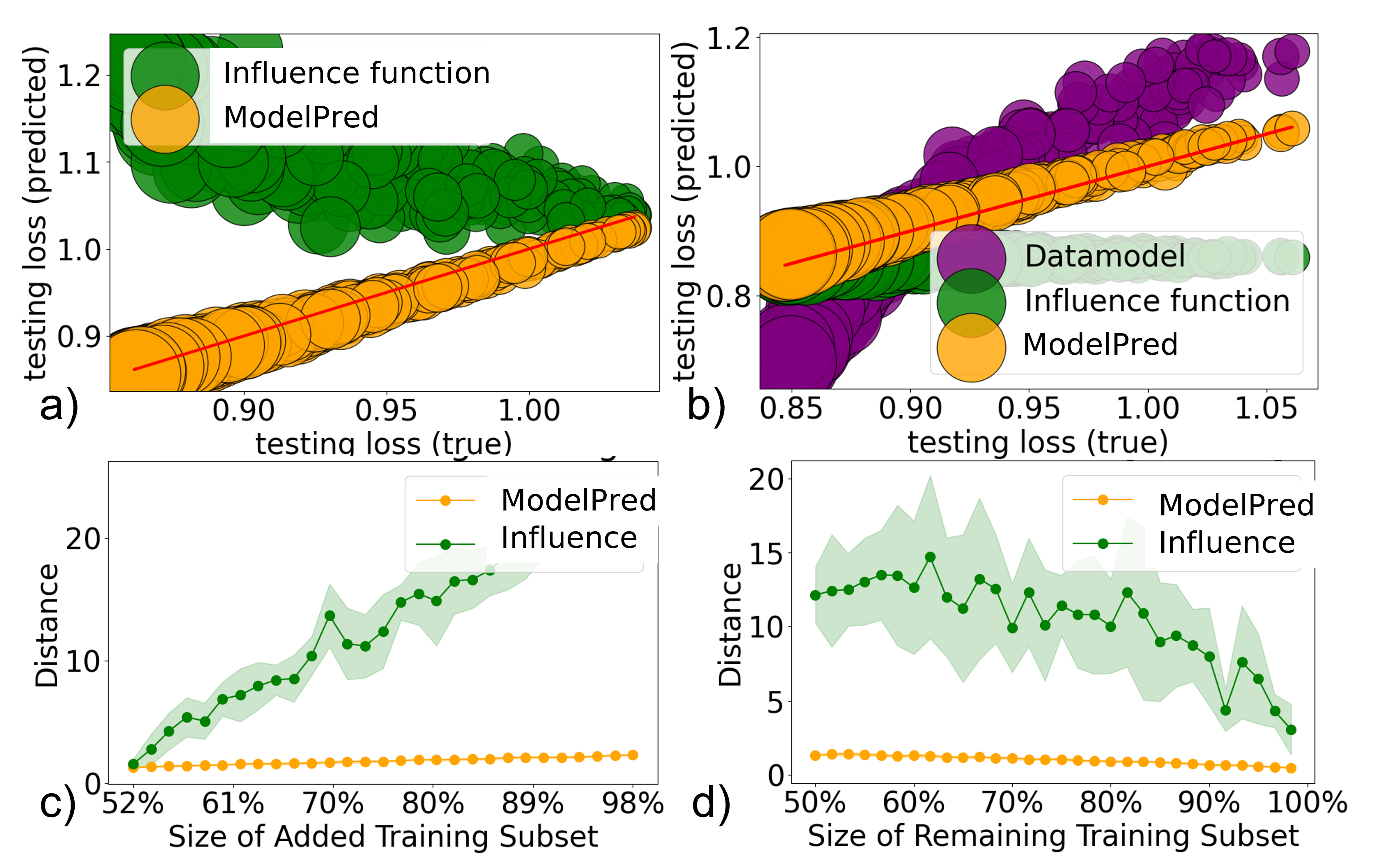}
\caption{\black{Results on MNIST dataset with LR as the base model. a) and b): the estimated loss by predicted parameters of each subset in the scenario of dataset deletion and addition; c) and d): the Euclidean distance between predicted parameters and optimal parameters in the same scenario.}}\label{fig:MNIST_LR}
\end{figure}

We visualize the result on MNIST in Figure~\ref{fig:MNIST_LR}.
The $x$ axis of Figure~\ref{fig:MNIST_LR} a) and b) is the ground-truth utility (\ie testing loss) and the $y$ axis shows the estimated utility.
The red line represents the ground-truth utility and each circle represents the estimated utility of one subset, where the circle size is proportional to the subset size.
It clearly shows that the error of loss predicted by Influence Function increases with the scale of the change on a dataset (either deletion or addition) since the utility estimated by Influence Function rapidly deviates from the red line, whereas \AlgName consistently provides accurate utility prediction.


Figure~\ref{fig:MNIST_LR} c) and d) demonstrate the Euclidean distance $\norm{\widehat{\theta}-\theta^*}$ with error bar where the $x$ axis in c) is the size of the training subset after addition and the $x$ axis in d) is the size of the remaining training subset after deletion, both of which are formatted as the percentage of the size of the full training set (\ie $n=300$ for MNIST). From Figure~\ref{fig:MNIST_LR} c) and d), it is shown that Influence Function can make an accurate approximation of the parameter change with a small scale of change (\ie at the beginning of data addition and deletion). However, the distance grows rapidly with the scale of change.
\black{The first-order Taylor approximation adopted in Influence Function cannot accurately estimate the effect of a large group of data points on the base model.
The inferior performance of Datamodel is also caused by its nature of linear approximation by linear regression, which cannot accommodate large dataset changes.}

For better readability, we defer the results of two scenarios with SVM as the base model to the Appendix, which demonstrate the similar performance of \AlgName in terms of efficiency and effectiveness.

By checking the performance of applying \AlgName  on transfer learning in NN.
similar conclusions  can be drawn compared to those of LR.
Firstly, \AlgName can accurately predict the high-dimensional parameters for the transfer learning of NN. 
Secondly, the utility of the refitted NN can be accurately estimated by the proposed method, which indicates the effectiveness and generalizability brought by the proposed regularizers.
Thirdly, the parameter prediction error is larger than the results of convex base models (\ie Table~\ref{table:lrdel} and ~\ref{table:lradd}), which might be due to the stochastic training process of NN.

\begin{figure}[ht]
\centering
\includegraphics[width=0.9\columnwidth]{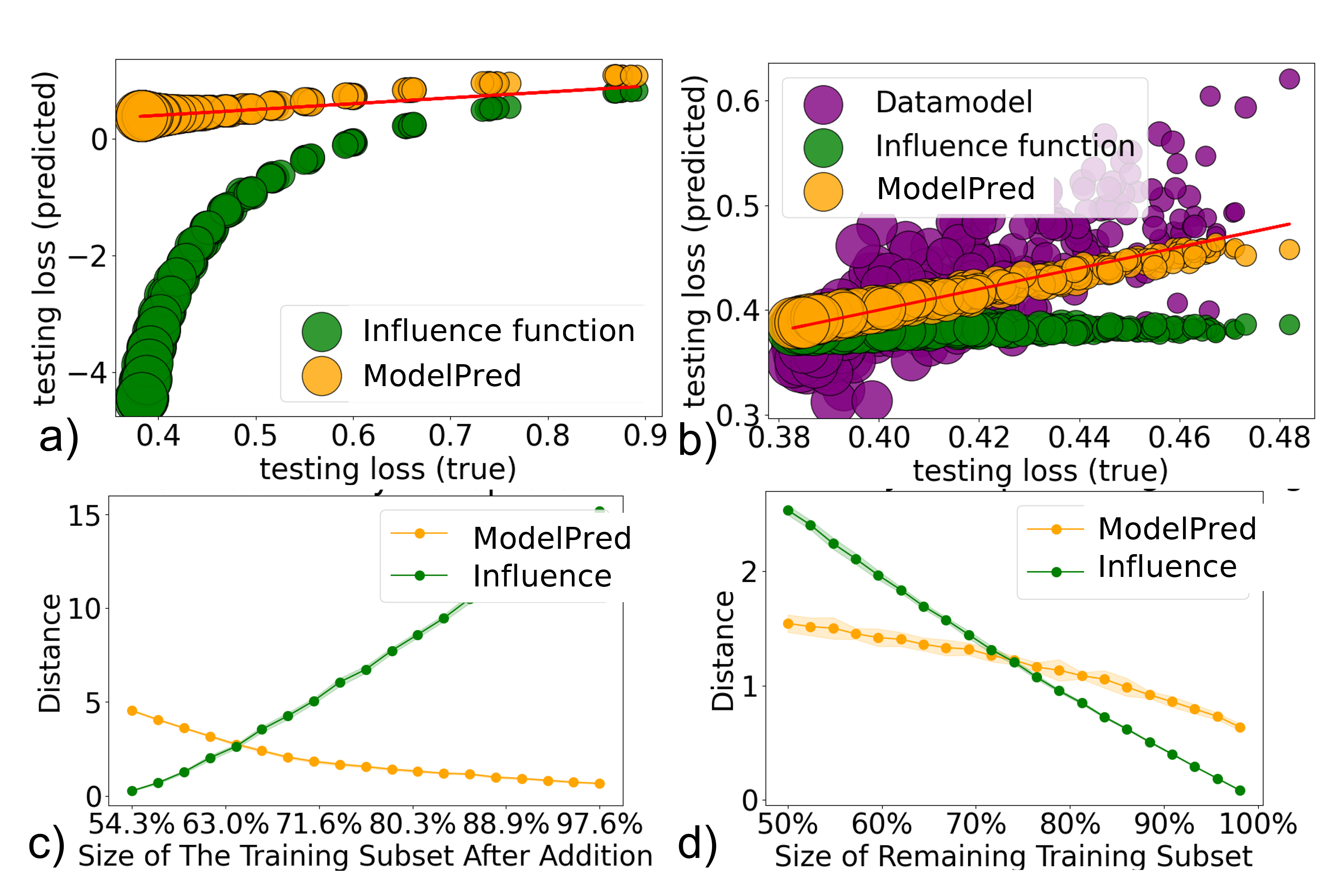}
\caption{\black{Results on Hymenoptera images with NN as the base model. a) and b): the estimated loss by predicted parameters of each subset in the scenario of dataset deletion and addition; c) and d): the Euclidean distance between predicted parameters and optimal parameters in the same scenario.}}\label{fig:hym_NN}
\end{figure}

Similar to Figure~\ref{fig:MNIST_LR}, Figure~\ref{fig:hym_NN} visualizes the estimated loss and the Euclidean distance between predicted parameters and optimal parameters by Influence Function and \AlgName with NN as the base model.
It shows the estimation error of the proposed method remains at a lower level after the intersection point.
This demonstrates the advantage of the proposed method over linear approximation by Influence Function, and it also quantifies the size of the subset after addition or deletion where the proposed method dominates the performance.
In summary, \AlgName demonstrates its superior performance in predicting the parameters for the training of ML models with subsets of different sizes.

\subsection{Applications}\label{sec:app}
We leverage \AlgName to a variety of ML applications to answer questions mentioned in Section~\ref{sec:intro}.\\

\emph{Which point is the most responsible for learning a given
parameter in the model?}

\textbf{$\AlgName$ Learns Learning Patterns}
To demonstrate the effectiveness of neural networks in learning the mappings between training data and the final parameters, we plot the saliency map of model parameter changes when a specific data point is excluded (``turned off'') from the training set. 
\black{In detail, LR is selected as the base model, and 300 data points from MNIST are used as the full training dataset.}
An LR model with L1 penalty \citep{tibshirani1996regression} is firstly trained on the full training set to identify the significant parameters. 
Then, a subset of parameters is randomly selected for visualization.
We flip the label of the first 20 samples in each training set to investigate the effect of noisy data on the fitted model, as well as to examine the generalizability of the proposed method.
Note that during the training process of \AlgName, we do not include the flipped samples.

\begin{figure*}[h]
\centering
\includegraphics[width=\textwidth]{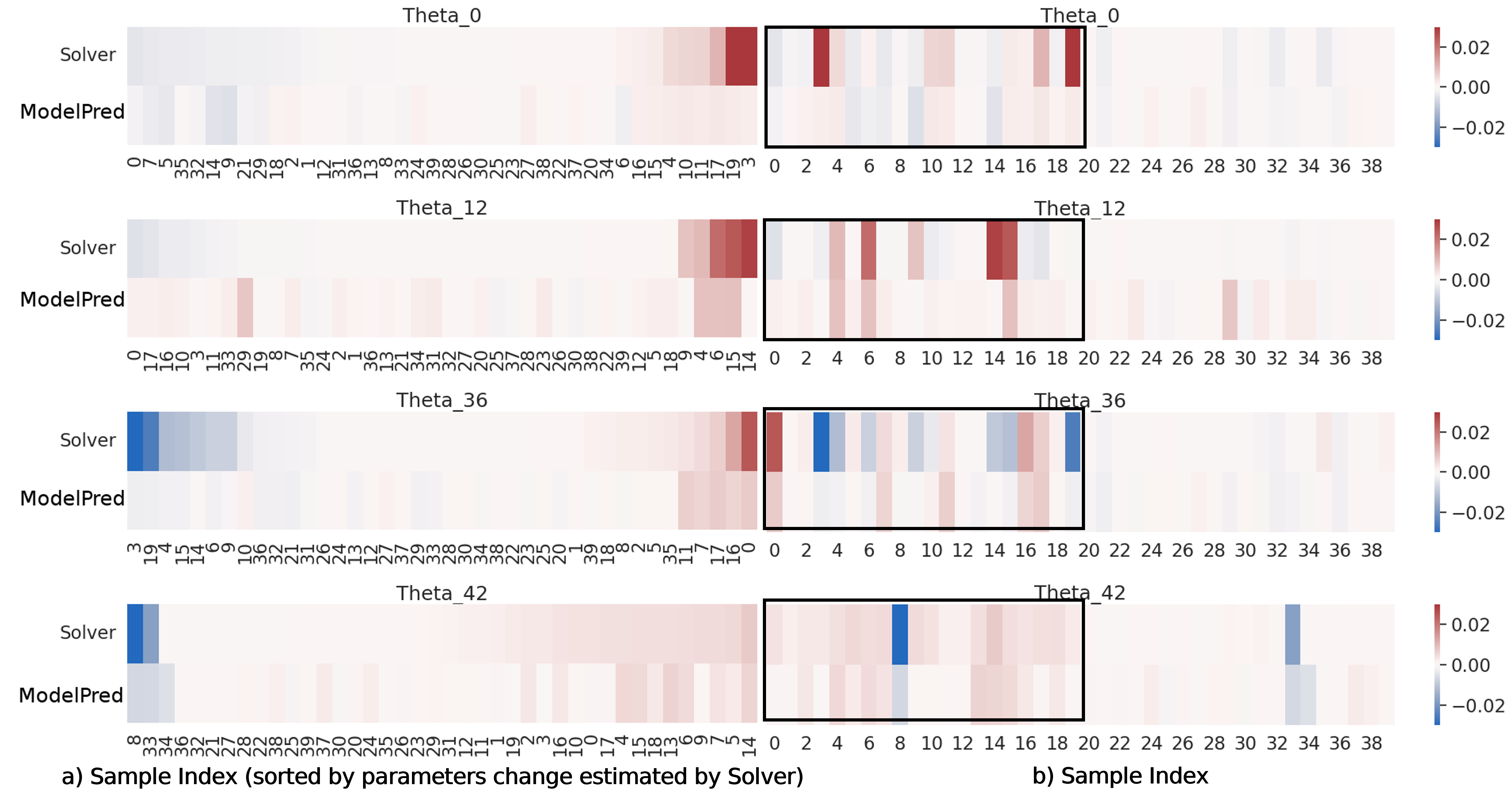}
\caption{\black{Saliency map of model parameter changes estimated by the solver and \AlgName with LR as the base model on MNIST dataset. a) Samples are ordered by the corresponding value of parameter changes estimated by the Solver. b) Samples are ordered by the original index. Flipped sample indexes are boxed.}}
\label{fig:Sailency_MNIST}
\end{figure*}
\black{By checking the second row of four parameters in the left panel of Figure  \ref{fig:Sailency_MNIST}, it shows the color of the reordered samples gradually changes from blue to red, which indicates \AlgName can accurately predict relative parameter change caused by removing individual samples.  
By comparing the color of the boxed samples on the right panel of Figure  \ref{fig:Sailency_MNIST}, the  parameter changes caused by noisy data (\ie first 20 samples) are more significant than others.}

\begin{table*}[]
\centering
\resizebox{0.85\textwidth}{!}{%
\begin{tabular}{c|cc|cc|cc|cc}
\toprule
\multirow{2}{*}{Method} & \multicolumn{2}{c}{Theta\_0} & \multicolumn{2}{c}{Theta\_12} & \multicolumn{2}{c}{Theta\_36} & \multicolumn{2}{c}{Theta\_42} \\\cmidrule{2-9}
 & noisy samples & good samples & noisy samples & good samples & noisy samples & good samples & noisy samples & good samples \\\midrule
Solver & -2.17E-03 & 5.27E-03 & -6.86E-03 & 4.70E-03 & -6.86E-03 & 3.32E-03 & -2.81E-03 & 3.74E-03 \\
Data2Model & -1.27E-03 & 9.68E-04 & 9.64E-04 & 1.66E-03 & -1.28E-03 & 1.47E-03 & -5.59E-04 & 2.28E-03\\
\bottomrule
\end{tabular}}
\caption{\black{Average parameter changes caused by the exclusion of noisy and good samples.}}
\label{table:pattern_MNIST}
\end{table*}
\black{We further investigate the average value of the parameter changes caused by the exclusion of noisy and good samples and summarize them in Table~\ref{table:pattern_MNIST}. 
By checking the average parameter change estimated by Solver, we find a systematic difference between noisy samples and good samples.
Removing noisy samples causes a decrease in the parameters while removing good samples leads to an increase. 
The change estimated by \AlgName is consistent with this pattern under most scenarios. 
This demonstrates the capability of \AlgName in accurately capturing the mapping from the training data to the learned model as well as its generalizability on noisy samples.}

\emph{What are the training points with the most or least contribution to the model?}

\textbf{Data Valuation.}
Quantifying the value of each training data point to a learning task is a fundamental problem in ML. SV is a widely used data value notion nowadays to identify the contribution of each training point \citep{ghorbani2019data,  jia2019efficient,jia2019towards, wang2020principled, jia2021scalability, wang2022data}. 
However, the exact SV calculation for a training dataset with $n$ points involves computing the marginal utility of every point in all subsets, which requires $2^n$ times of utility evaluation by retraining the model.
Since \AlgName is capable of efficient utility evaluation by predicting the model parameters on a new training subset, \AlgName can speed up the SV calculation.

To validate the performance of \AlgName on calculating SV, we use permutation sampling \citep{maleki2015addressing} to generate the subsets evaluated from the full training set $D$ in sequence.
We compare the SV results calculated by \AlgName, and \textbf{UtlLearn}\xspace (\ie predicted SV) with the SV calculated by the solver as ground truth, and we also record the total time consumed by each method for subset utility evaluation.
Here, we create \textbf{UtlLearn} as a baseline which has the same neural network structure as \AlgName but directly predicts the utility of a subset, and thus, not including KKT loss.
We set the number of permutations $T\in \{10, 50, 100, 500, 1000\}$ for Iris, SPAM, and HIGGS.
For MNIST, we set $T\in \{10, 50, 100, 200\}$  due to the long computation time of the solver caused by high feature dimension.
Similarly, we use NRMSE and Spearman correlation as performance metrics to evaluate the predicted SV.

\begin{table*}
\resizebox{\textwidth}{!}{%
\begin{tabular}{cc|ccc|ccc|ccc|ccc}
\toprule
\multicolumn{2}{c|}{Permutation} &
  \multicolumn{3}{c|}{50} &
  \multicolumn{3}{c|}{100} &
  \multicolumn{3}{c|}{500} &
  \multicolumn{3}{c}{1000} \\ 
\multicolumn{2}{c|}{} &
  \multicolumn{2}{c}{Shapley value} &
  \multirow{2}{*}{Time (sec)} &
  \multicolumn{2}{c}{Shapley value} &
  \multirow{2}{*}{Time (sec)} &
  \multicolumn{2}{c}{Shapley value} &
  \multirow{2}{*}{Time (sec)} &
  \multicolumn{2}{c}{Shapley value} &
  \multirow{2}{*}{Time (sec)} \\ \cmidrule{3-4} \cmidrule{6-7} \cmidrule{9-10} \cmidrule{12-13}
Dataset &
  Algorithm &
  NRMSE &
  Spearman &
   &
  NRMSE &
  Spearman &
   &
  NRMSE &
  Spearman &
   &
  NRMSE &
  Spearman &
   \\ \midrule
\multirow{3}{*}{Iris} &
  \textbf{\AlgName} &
  \textbf{8.52\%} &
  \textbf{0.9265} &
  \textbf{1.05E+01} &
  \textbf{7.69\%} &
  \textbf{0.9463} &
  2.22E+01 &
  \textbf{6.82\%} &
  \textbf{0.9747} &
  \textbf{7.84E+01} &
  \textbf{6.22\%} &
  \textbf{0.9585} &
  2.27E+02 \\ 
 &
  UtlLearn &
  25.97\% &
  -0.2526 &
  1.08E+01 &
  28.46\% &
  -0.1118 &
  \textbf{2.21E+01} &
  36.14\% &
  0.0388 &
  7.87E+01 &
  42.79\% &
  -0.0480 &
  \textbf{2.22E+02} \\ 
 &
  solver & --
   & --
   &
  4.05E+01 & --
   & --
   &
  8.64E+01 & --
   & --
   & 
  2.70E+02 & --
   & --
   & 8.57E+02 \\ \midrule
\multirow{3}{*}{SPAM} &
  \textbf{\AlgName} &
  \textbf{13.46\%} &
  \textbf{0.3774} &
  \textbf{1.00E+02} &
  \textbf{18.16\%} &
  \textbf{0.3686} &
  \textbf{1.23E+02} &
  \textbf{15.57\%} &
  \textbf{0.7069} &
  1.43E+02 &
  \textbf{18.51\%} &
  \textbf{0.7772} &
  2.86E+02 \\ 
 &
  UtlLearn &
  15.21\% &
  -0.2081 &
  1.14E+02 &
  20.76\% &
  -0.2091 &
  1.26E+02 &
  18.72\% &
  -0.1187 &
  \textbf{1.42E+02} &
  22.96\% &
  -0.0600 &
  \textbf{2.75E+02} \\ 
 &
  solver & --
   & --
   &
  3.07E+02 & --
   & --
   & 
  6.50E+02 & --
   & --
   & 
  8.96E+02 & --
   & --
   &
  4.73E+03 \\ \midrule
\multirow{3}{*}{HIGGS} &
  \textbf{\AlgName} &
  \textbf{12.15\%} &
  \textbf{0.5249} &
  1.37E+01 &
  \textbf{15.60\%} &
  \textbf{0.4463} &
  \textbf{2.42E+01} &
  \textbf{17.54\%} &
  \textbf{0.6655} &
  \textbf{8.30E+01} &
  \textbf{17.45\%} &
  \textbf{0.7066} &
  \textbf{1.69E+02} \\ 
 &
  UtlLearn &
  43.12\% &
  -0.2494 &
  \textbf{1.27E+01} &
  61.82\% &
  -0.3029 &
  2.15E+01 &
  86.85\% &
  -0.5729 &
  1.02E+02 &
  90.52\% &
  -0.6742 &
  1.75E+02 \\ 
 &
  solver & --
   & --
   &
  9.17E+01 & --
   & --
   &
  1.70E+02 & --
   & --
   &
  6.39E+02 & --
   & --
   & 
  1.27E+03 \\ \bottomrule
\end{tabular}
}
\caption{A summary of \AlgName results and baseline comparison results in SV prediction with LR on IRIS, SPAM, and HIGGS with permutation number $T \in \{50, 100, 500, 1000\}$. Mean and standard deviation reported over 10 experimental trials. The best result is highlighted in \textbf{bold}.}
\label{table:shap3}
\end{table*}

Table~\ref{table:shap3} summarizes the results for Iris, SPAM, and HIGGS.
It can be observed that \AlgName outperforms \textbf{UtlLearn} in terms of SV prediction under all scenarios.
Compared to \textbf{UtlLearn}, \AlgName has more network parameters to train due to the higher output dimension $\dparam$.
However, \AlgName gains its advantage from the KKT regularization which enforces the optimality of predicted parameters, thus, achieving accurate parameter and utility prediction.

Additionally, comparing the results of a varied number of permutations, it is shown that the Spearman correlation tends to increase with the number of permutations $T$.
When $T$ is small, the ground-truth SV calculated by permutation sampling is very sensitive to the utility of small subsets.
Therefore, the relatively inaccurate prediction on small subsets results in the high NRMSE and low Spearman correlation.
With the increase of $T$, the SV predicted by \AlgName can better represent the true SV with the increasing correlation.

\begin{figure}[h]
\centering
\includegraphics[width=0.9\columnwidth]{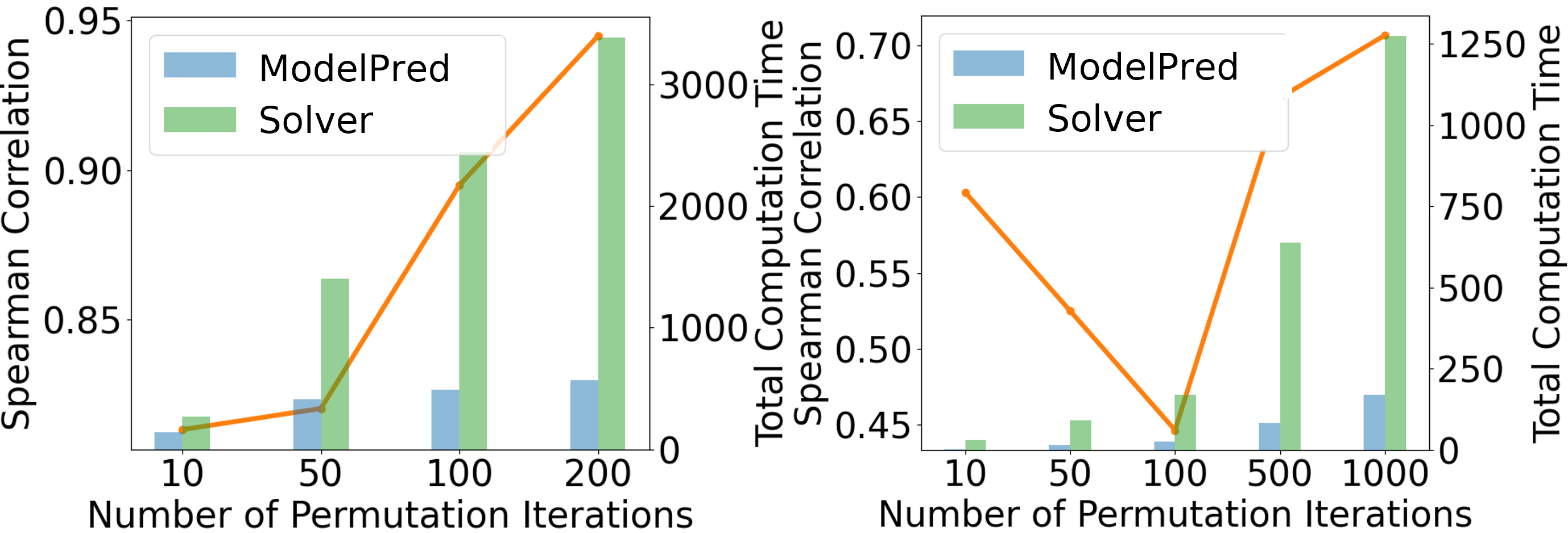}
\caption{Total computation time consumed by \AlgName and the solver on training a different number of subsets in SV calculation for MNIST (left) and HIGGS (right).}
\label{fig:MNIST_shapley}
\end{figure}

Figure~\ref{fig:MNIST_shapley} presents the trends of Spearman Correlation of predicted SV by \AlgName, and compares the total computation time of SV calculation with the solver on MNIST and HIGGS. 
It demonstrates the advantage of \AlgName over the solver in computation time, where time consumed by \AlgName has a slow growth with the increasing number of permutations.

\emph{How to select data that benefit model performance?}

\textbf{Data Selection.}
Despite the rapid growth of big data, the performance of all learning algorithms is upper bounded by the quality of training data \citep{jain2020overview}.
Low-quality training data could be attributed to the contamination of various harmful examples (\ie noisy samples, mislabeled samples ) or the lack of representativeness of the data distribution.
Furthermore, a small but high-quality training dataset can significantly reduce the computation workload of an ML model as well as maintain comparable utility. 

\begin{figure}[h]
\centering
\includegraphics[width=0.9\columnwidth]{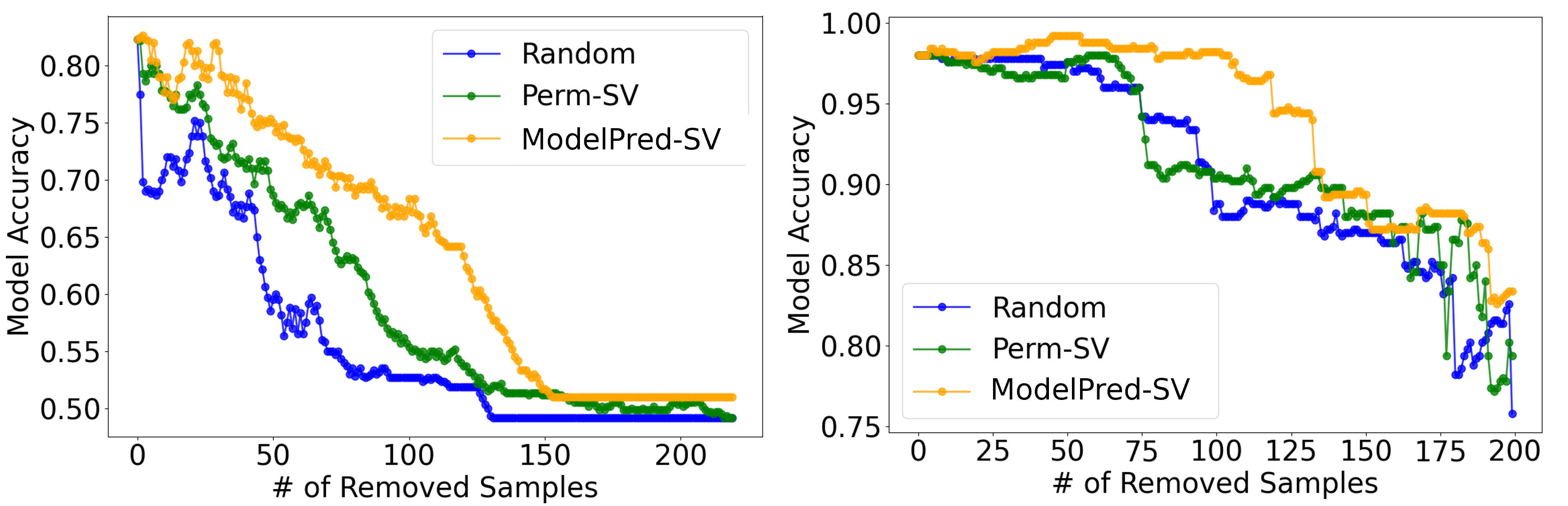}
\caption{Testing accuracy with LR retrained on the remaining training samples for SPAM (left) and MNIST dataset (right).}
\label{fig:SPAM_MNIST_Bad}
\end{figure}

In this application, we aim to efficiently identify the quality of training data by estimating the Shapley value of each data point.
We estimate the SV of training samples in SPAM and MNIST by \AlgName with LR as the base model (\ie \AlgName-SV).
The label of 10\% training samples is flipped in the experiment.  
Then, the samples are removed from the training set according to their SV from the smallest to the largest.
The base model is retrained on the shrunk training set to obtain classification accuracy on the testing set. 
Shapley value estimated by Permutation Sampling with L-BFGS-B solver (\ie Perm-SV), and randomly selecting the sample to be removed (\ie Random) are compared with \AlgName as baselines.
To maintain similar computation time by \AlgName-SV and Perm-SV, 1000 and 50 permutations are performed to calculate SV, respectively.
Figure~\ref{fig:SPAM_MNIST_Bad} shows that after removing the sample number of samples, \AlgName maintains a higher testing accuracy than the baselines while Perm-SV outperforms Random in most cases.
This indicates that SV can effectively identify the quality of a data point, and \AlgName-SV can estimate the SV more accurately with a larger number of permutations.
Figure~\ref{fig:SPAM_MNIST_Bad} also shows that there is an increasing trend of the testing accuracy on both SPAM and MNIST in the initial phase by \AlgName-SV, which indicates that the mislabeled low-quality data are removed effectively to improve the model's learning performance.

\emph{Is a training data point memorized by the model? }

\textbf{Memorization and Influence Score Estimation.}
One of the most important features of ML algorithms is their capability for generalization. 
Label memorization score has been proposed to identify and understand the utility of typical, atypical samples and outliers for a model in order to understand the generalization \citep{feldman2020neural}.
For a model $f(x;\theta)$ trained on a dataset $D$ by a learning algorithm $\A$, the label memorization score on sample $(x_i, y_i) \in D$ is defined as:
\begin{align}
    \label{eq:memorization}
    \operatorname{mem}(\A, D, i) &:= 
    \operatorname{Pr}_{f \leftarrow \mathcal{A}(D)}\left[f\left(x_{i}\right)=y_{i}\right]- \\\nonumber
    &\operatorname{Pr}_{f \leftarrow \mathcal{A}(D \backslash i)}\left[f\left(x_{i}\right)=y_{i}\right],
\end{align}
where $D\backslash i$ denotes the subset $D$ with $(x_i, y_i)$ removed.

Following \citep{feldman2020neural}, we adopt the \emph{subsampling-based estimation algorithm} to estimate the memorization score. 
Specifically, we sample $m$ subsets of equal size $|S_i| = 0.7|D|$ and the memorization score can be estimated as the following:
\begin{align}
    \widehat{\mem}&(\A, D, i) = \nonumber \\
    &\frac{1}{|\{j: x_i \in S_j\}|} \sum_{j: x_i \in S_j} I[f\left(\A\left(S_{j}\right), x_i\right) = y_i] \nonumber \\ \nonumber
    &- \frac{1}{|\{j: x_i \not\in S_j\}|} \sum_{j: x_i \not\in S_j} I[f\left(\A\left(S_{j}\right), x_i\right) = y_i].
\end{align}


However, with a large dataset $D$, the number of subsets $m$ is substantially large to accurately estimate the label memorization of each sample \citep{feldman2020neural}. 
This computation constraint can be partially overcome by the \AlgName since it can efficiently estimate the trained model on a subset, thus capable of accelerating the memorization score estimation.

\begin{figure}[h]
\centering
\includegraphics[width=\columnwidth]{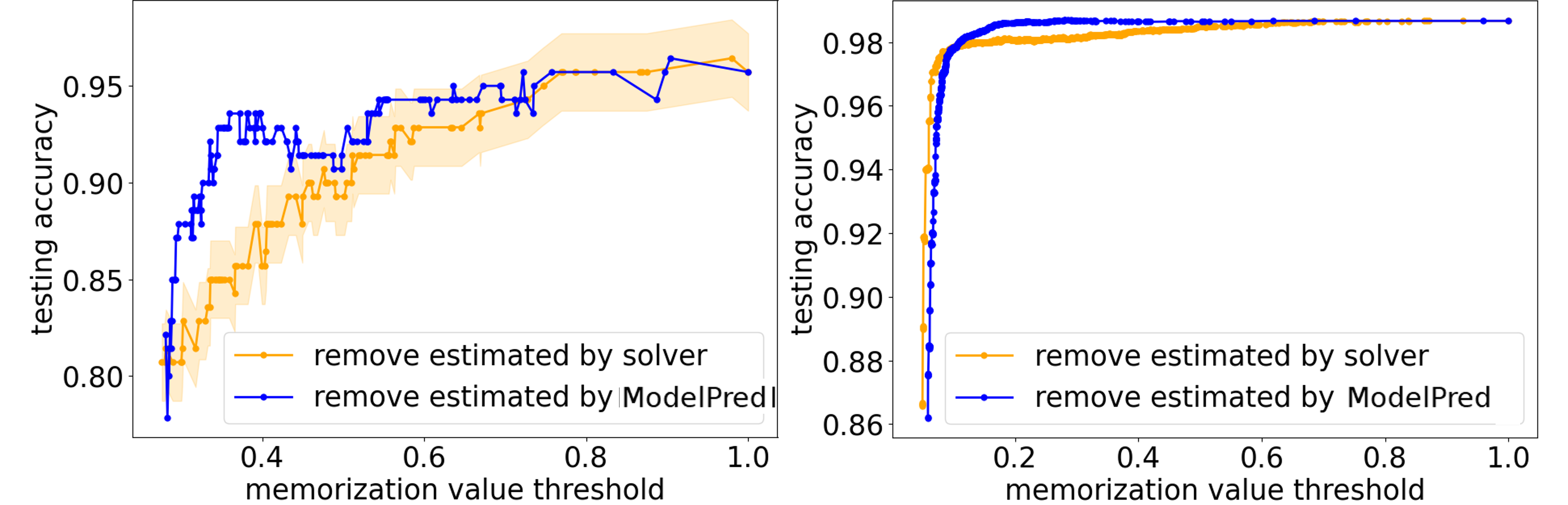}
\caption{Testing accuracy after removing samples larger than the memorization value threshold (\ie ground-truth and estimated by \AlgName) on Hymenoptera with pre-trained NN (left) and on MNIST with LR (right). Mean and standard deviation reported over 10 experimental trials.}
\label{fig:memory_acc}
\end{figure}

In the experiment, two base models (NN and LR) with two datasets (Hymenoptera and MNIST) are employed for memorization estimation. 
As mentioned in Section~\ref{subsec:exp}, the NN is pre-trained on CIFAR-10 and the last layer is trained to adjust to Hymenoptera.
In order to maintain comparable computational time for the comparison between \AlgName and the SGD solver, 1000 subsets are randomly generated for each sample with a size of 280 
for \AlgName, and 50 subsets are generated for the SGD solver to estimate the memorization of 208 training samples in Hymenoptera. 
In order to demonstrate and compare the estimation accuracy of memorization scores, we depict the effect of removing memorized samples on test accuracy. 
Specifically, in the left panel of Figure \ref{fig:memory_acc}, we show the model test accuracy when data points whose memorization scores are higher than a certain threshold are excluded from the training set. 
As we can see, memorization scores estimated by \AlgName result in higher test accuracy compared with the baseline estimator, which means that \AlgName can improve memorization score estimation and thus better identify outliers.



When adopting LR as the base model and MNIST as the dataset, memorization scores of the first 200 samples are estimated by  L-BFGS-B (\ie the solver for LR) and \AlgName on 500 testing samples. 
Since the learning algorithm, in this case, is deterministic, we modify the definition of memorization score as the following:
\begin{align}\label{eq:memorization2} %
    &\mem(\A, D, i) \\\nonumber
    &:= \conf(f_{\mathrm{include}}(x_i), y_i) - \conf(f_{\mathrm{exclude}}(x_i), y_i),
\end{align}
where $f_{\mathrm{include}} = \A(D)$ and $f_{\mathrm{exclude}} = \A(D \setminus i)$, and $\conf(f, y)$ denote the confidence score of $f$ on class $y$. 


\begin{figure}[h]
\centering
\includegraphics[width=0.6\columnwidth]{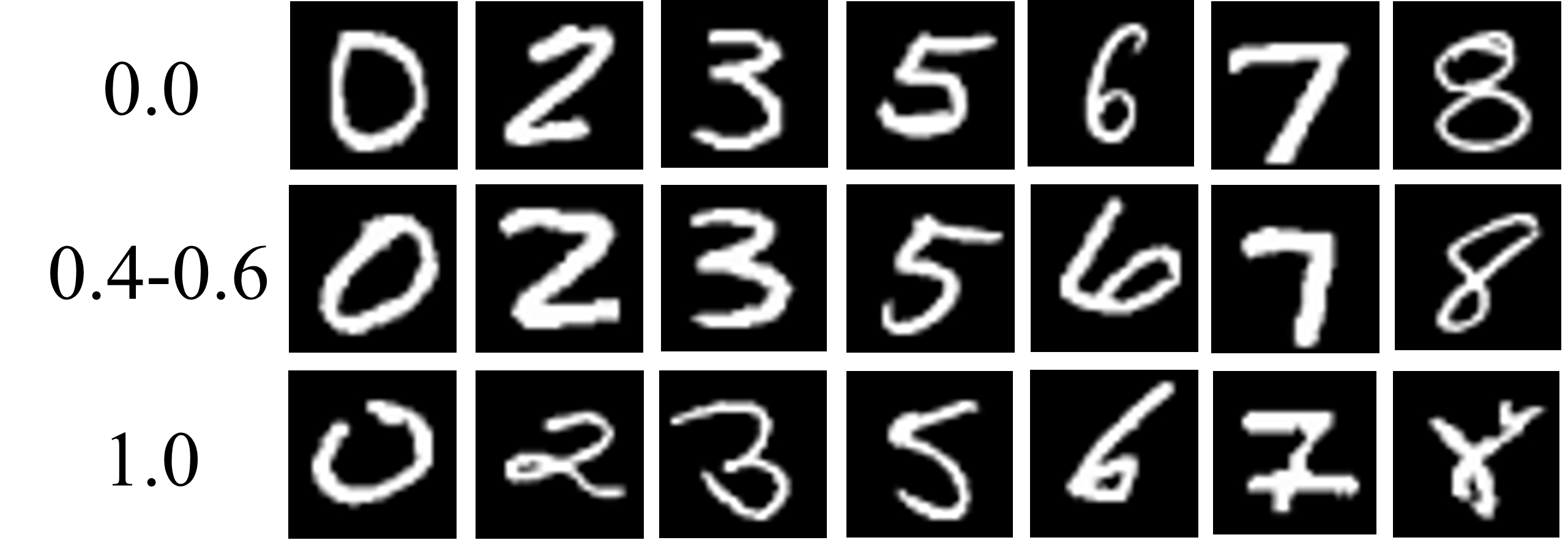}
\caption{Examples of memorization values estimated by \AlgName from the first 200 samples of MNIST class 0, 2, 3, 5, 6, 7, 8.}
\label{fig:MNIST_memo_example}
\end{figure}

The right panel of Figure~\ref{fig:memory_acc} demonstrates the effect of removing samples with the memorization values estimated by \AlgName and the solver, which can be treated as the ground-truth memorization value,  on the test accuracy of MNIST dataset. 
As shown by Figure~\ref{fig:memory_acc}, the trend of the removal effect estimated by \AlgName is very close to the ground-truth trend.
Compared to the solver, it takes 1.84\% of the time to estimate the value by \AlgName, which greatly reduces the computation workload.


Figure~\ref{fig:MNIST_memo_example} visualizes samples with memorization values estimated by \AlgName around 0, 0.5, and 1, respectively, which shows the effectiveness of \AlgName in identifying memorized examples.  
Intuitively, samples with an estimated memorization value of 0 are relatively normal, whereas those with a value of 1 are atypical (e.g., highly unclear or incorrectly categorized). Therefore, samples with high memorization values should be carefully validated to be included in the training dataset. 

\begin{figure}[h]
\centering
\includegraphics[width=0.6\columnwidth]{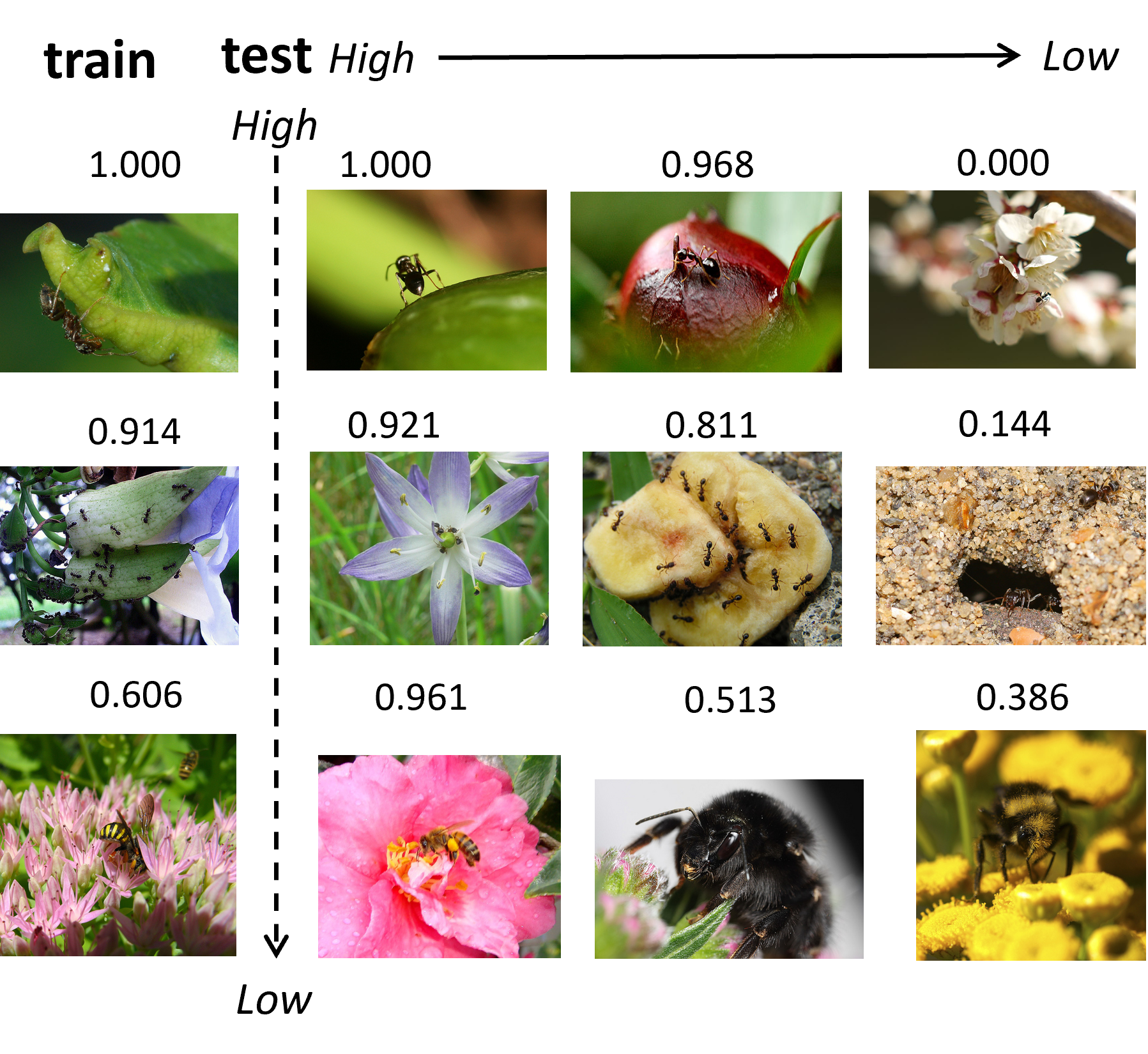}
\caption{Examples of selected influence pairs in Hymenoptera. The left column shows the examples with memorization from high to low in the training set. For each training example, examples with high and low estimates are presented in the testing set.}
\label{fig:high_influ_Hym}
\end{figure}

\begin{figure}[h]
\centering
\includegraphics[width=0.6\columnwidth]{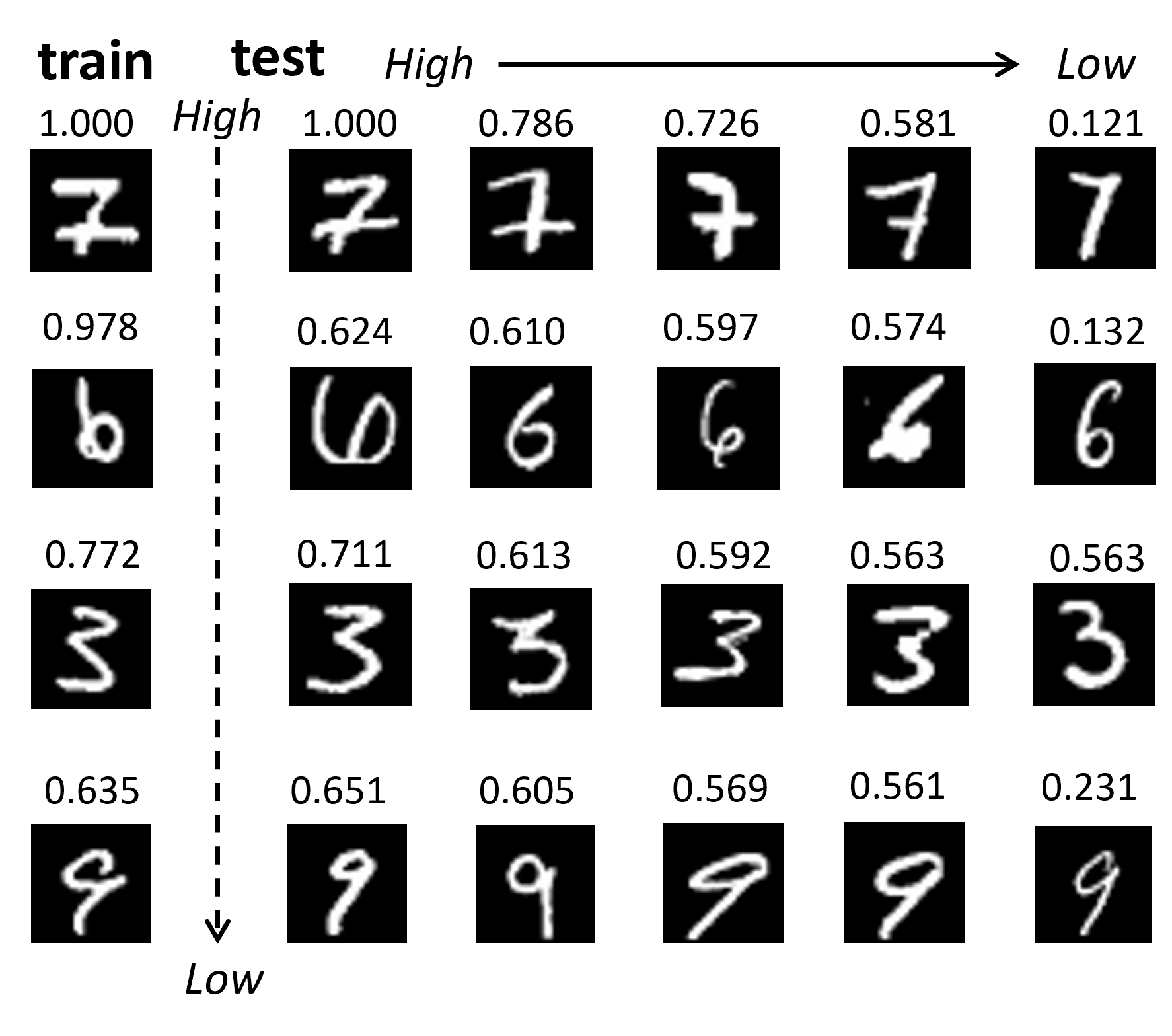}
\caption{Examples of selected influence pairs in MNIST. The left column shows examples of memorization score from high to low in the training set. For each training example, examples with high and low estimates are presented in the testing set.}
\label{fig:high_influ_MNIST}
\end{figure}

Furthermore, we present pairs of examples with influence estimated by \AlgName from high to low in Hymenoptera and MNIST as shown in Figure~\ref{fig:high_influ_Hym} and \ref{fig:high_influ_MNIST}.
The influence of a training sample $(x_i, y_i)$ on a testing sample $z=(x,y)$ is measured as:
\begin{align}
 \operatorname{infl}(\mathcal{A}, D, i, z):& =\operatorname{Pr}_{f \leftarrow \mathcal{A}(D)}[f(x)=y]-\\\nonumber
  & \operatorname{Pr}_{f \leftarrow \mathcal{A}(D\backslash i)}[f(x)=y].
\end{align}
As shown in Figure~\ref{fig:high_influ_Hym} and \ref{fig:high_influ_MNIST}, high-influence pairs correspond to similar (or even near-duplicated) images.
These samples in the test set benefit mostly from the high-memorized training samples due to the long-tail distribution.
In contrast, examples in pairs with lower influences are more regular.
This validates the effectiveness of the influence memorization value estimated by the proposed \AlgName, which helps to better interpret the influence of samples and the generalization of ML algorithms.

\newcommand{\acc}{\operatorname{acc}}

\emph{How to produce accurate estimates of uncertainty on model predictions?}

\textbf{Model Calibration.} 
Classification models' confidence calibration performance is crucial in mission-critical tasks \citep{guo2017calibration}, and a well-calibrated model should have a confidence (\ie probability associated with the predicted class label) matching with its ground truth accuracy.
Expected Calibration Error (ECE) \citep{naeini2015obtaining} is the primary metric of model calibration performance, which is calculated by partitioning predictions (with range [0, 1]) into $M$ equally-spaced bins and calculating the weighted average of differences between bins' accuracy and confidence:
\[\mathrm{ECE}=\sum_{m=1}^{M} \frac{\left|B_{m}\right|}{n
}\left|\acc\left(B_{m}\right)-\conf\left(B_{m}\right)\right|,\]
where $n$ is the number of training samples, $B_m$ is the set of indices of samples whose prediction confidence falls into the interval $I_m = (\frac{m-1}{M},\frac{m}{M})$ for m $\in$ \{1,\dots, M\}. Accuracy and confidence are calculated below:
\begin{align}
    \acc(B_{m}) &= \frac{1}{|B_{m}|} \sum_{i \in B_{m}} \mathbbm{1}[\widehat{y}_{i}=y_{i}],\\
    \conf(B_{m}) &= \frac{1}{|B_{m}|} \sum_{i \in B_{m}} \widehat{p}_{i},
\end{align}
where $\widehat{y}_{i}$ and $y_i$ are the predicted and true class labels of sample $i$, respectively, and $\widehat{p}_i$ is the confidence for sample $i$ on label $y_i$.
A model with better calibration performance should have a lower ECE.

\begin{table}[h]
\centering
\resizebox{0.5\columnwidth}{!}{%
\begin{tabular}{l|l|c}

Dataset & Regular & \multicolumn{1}{c}{\textbf{\AlgName}} \\ \midrule
Iris    & 0.0562  & \textbf{0.0508 ± 0.0002}               \\ 
SPAM    & 0.2693  & \textbf{0.1661 ± 0.0054}               \\ 
MNIST   & 0.5492  & \textbf{0.5227 ± 0.0014}               \\ 
\end{tabular}
}
\caption{Expected Calibration Error (ECE). `Regular' refers to an ensemble of LR models obtained during the training of $\Adnn$; `\AlgName' refers to an ensemble of a combination of these models and 5000 models generated by the trained $\Adnn$.}
\label{table:ece}
\end{table}

One way to improve a model's calibration performance is bagging \citep{breiman1996bagging}, which generates an ensemble of models on subsampled datasets and aggregates over each prediction to obtain final prediction results. 
However, training a large number of models can be time-consuming. 
We show here that \AlgName provides an efficient way to perform bagging, which effectively reduces the ECE. 
Specifically, we use LR base models as a baseline of the ensemble. 
Further, we combine these models with models predicted by the trained DNN, $\Adnn$, to get a larger ensemble. 
To generate models using $\Adnn$, each time we randomly sample a subset of instances with a ratio $0.6$ from the training set with replacement and obtain estimated models predicted by $\Adnn$. 
As shown in Table~\ref{table:ece}, combining more models generated by $\Adnn$ effectively lower the ECE on Iris, SPAM, and MNIST dataset. 
We exclude the HIGGS dataset in this application due to the low performance of the base model, LR, on this dataset, which results in a large variance in the prediction.
Additionally, temperature scaling \citep{guo2017calibration} might further improve the model's calibration performance and we consider investigating the combination of temperature scaling and our method with model ensemble as the future work.



\section{Related Work}\label{sec:related}
\textbf{Rapid Model Parameter Approximation.} Machine unlearning and incremental model maintenance provide post-hoc techniques to estimate model parameters without retraining from scratch. \citet{ginart2019making, nguyen2020variational, brophy2021machine} have investigated unlearning strategies on specific types of learning algorithms, whereas \citet{bourtoule2021machine} provides generally applicable strategies. 
For simple linear models, \citet{cauwenberghs2001incremental, schelter2019amnesia, wu2020priu} introduce incremental model maintenance techniques for efficient updating a model for both data deletion and addition. 
In DNNs, Influence Function \citep{koh2017understanding} is proposed to measure the effect of a manipulated data point by using Taylor expansion to approximate model parameters. 
DeltaGrad \citep{wu2020deltagrad} saves optimizer’s update steps in order to more accurately approximate the removal of multiple sample points. 
\citet{golatkar2020eternal} used the approximation of the Fisher Information Matrix for the remaining data sample to measure the update step to modify the model parameters. 
Although all techniques are capable of efficient model parameter approximation for a change with a small number of samples \citep{basu2020influence, mahadevan2021certifiable}, they are not scalable for situations when a large number of training points are altered.

\textbf{Learning to Optimize.} L2O focuses on learning the optimization algorithm (\ie optimizer). 
An early approach was provided in \citet{andrychowicz2016learning} to use Recurrent Neural Networks (RNN) to learn the optimizer. 
\black{For larger model training, it requires the RNN model to iterate through more time steps. Unfortunately, that will create a vanishing gradient or an exploding gradient of the RNN optimization, which in turn will render unstable training of L2O. Due to these barriers, more works \citep{chen2017learning, lv2017learning, wichrowska2017learned, metz2019understanding, cao2019learning, chen2020rna} focus on overcoming those problems by improving the LSTM structure. 
Even though, those works focus mainly on specific optimizers' family, such as SGD, Adam, or RMSProp. 
As another branch, \citet{li2016learning} proposed a reinforcement learning (RL) technique, in which the RL policy is the update step of the optimizer and the reward is the loss for the optimizer. 
However, RL methods are not scalable. 
Therefore, \citet{almeida2021generalizable}
proposed to update the optimizer’s hyperparameters instead of learning to update parameters, which helps to improve the generalization ability of L2O models.}
Despite the advantages of fast optimization and potential generalizability, L2O learns to learn and optimize the optimizer, which is computationally intensive to be applied to repeated model training on a large number of subsets.

\textbf{Learning Optimization by DNNs.} 
A line of research studies using deep learning to solve convex optimization problems by encoding constraints and dependencies into network structure to find the optimal end-to-end mapping. 
\citet{agrawal2019differentiable} embedded disciplined convex optimization problems as differentiable layers within DNN architectures as a new solver.
\citet{amos2017optnet} introduced a network architecture to solve differentiable optimization by end-to-end deep learning training.
Similar to L2o, repeated training is still computationally demanding with these approaches. 
Moreover, they also suffer from poor generalization performances, as they are limited to linear programs (LP), and quadratic programs, and are difficult to be applied to other settings. 
\citet{chen2020learning} proposed to use DNN to predict the optimal set of active constraints to improve the generalizability, but only for the form of the linear program.
\black{\citet{carlini2022membership} trained a set of models on the subset of the training dataset to perform the membership inference attack on a target model. 
However, they did not use such a technique to understand the input-output behavior of a learning algorithm, which is the focus of our work.}








\section{Conclusion}\label{sec:clu}
We propose \AlgName, as a framework to analyze the dependence of the trained models on training data via supervised learning. We introduce two novel regularization techniques to prevent overfitting the context of predicting models from training data.
We show that the expressiveness of DNNs makes them suitable for approximating the input-output behavior of a learning algorithm. 
We showcase the applications of \AlgName to build trust in ML.

For future work, it is intriguing to rigorously study the learnability of the mapping from data to the trained model using neural networks. Moreover, while providing promising results, our current design simply examines the $\ell_2$ distance between the ground-truth and the predicted model parameters. It is interesting to explore more sophisticated ways to measure parameter distance that accounts for the difference in predictive behaviors of the two models. 

\newpage

\section*{Acknowledgement}
This work is supported by Princeton’s Gordon Y. S. Wu Fellowship and Cisco Research Awards, Amazon-Virginia Tech Initiative in Efficient and Robust Machine Learning.
We thank Tong Wu for the advice on implementation.

\bibliographystyle{plainnat}
\bibliography{ref.bib}

\newpage
\onecolumn

\appendix


\section{Proofs}

\subsection{Theorem 1}

\begin{customthm}{1}[Restated]
If for all $z \in [0, 1]^{d}$, the loss function $\ell(\theta; z)$ is $\alpha$-strongly convex in $\theta$, and $\norm{\frac{\del}{\del \theta \del z_i} \ell (\theta, z)} \le \bound$ for $i \in [d]$, 
then for $\theta_k (D) = [\argmin_\theta L(\theta; D)]_k$, where $L(\theta; D) = \frac{1}{n} \sum_{i=1}^n \ell(\theta; (x_i, y_i)) + \frac{\lambda}{2} \norm{\theta}_2^2$ and $[\cdot]_k$ means the $k$th element in the vector, then we have 
\begin{align}
    \norm{\frac{\del \theta_k}{\del D}}
    \le \bound \frac{\sqrt{\dparam (d+1)}}{\sqrt{n} (\alpha + \lambda)}.
\end{align}
\end{customthm}
\begin{proof}
Given a loss function $L(\theta; D) = \frac{1}{n} \sum_{i=1}^n \ell(\theta; (x_i, y_i)) + \frac{\lambda}{2} \norm{\theta}_2^2$, the unique local minimum $\widehat \theta(D)$ is characterized by the implicit function 
\begin{align}
    \g L(\widehat \theta, D) 
    = \frac{1}{n} \sum_{i=1}^n \g_\theta \ell(\widehat \theta, (x_i, y_i)) + \lambda \widehat \theta = 0.
\end{align}

Denote $\sigma(j) = \lceil j / (d+1) \rceil$ to be the index of the data point (and hence the loss function $\ell$) $j$th dimension of $D$ corresponds to. By the implicit function theorem and the chain rule, we have 
\begin{align}
     \bigg(\sum_{i=1}^n\frac{\del \ell_i}{\del \theta^T \del \theta} + n \lambda I\bigg)\frac{\del \theta}{\del D_j} + \frac{\del \ell_{\sigma(j)}}{\del \theta \del D_j} = 0,
\end{align}
where $\ell_i = \ell(\cdot; (x_i, y_i))$ and, therefore, we obtain 
\begin{align}
    \frac{\del \theta}{\del D_j} = -[H^{-1}]\frac{\del \ell_{\sigma(j)}}{\del \theta \del D_j},
\end{align}
where $H = \sum_{i=1}^n \frac{\del \ell_i}{\del \theta^T \del \theta} + n\lambda I $ is the Hessian matrix. We use $[H^{-1}]_k$ to denote the $k$th row of the Hessian inverse, and, hence, 
\begin{align}
    & \frac{\del \theta}{\del D_j} = \left[ \frac{\del \theta_1}{\del D_j}, \ldots,  \frac{\del \theta_{\dparam}}{\del D_j} \right] \\
    &= \left[ -[H^{-1}]_1\frac{\del \ell_{\sigma(j)}}{\del \theta \del D_j}, \ldots, -[H^{-1}]_{\dparam}\frac{\del \ell_{\sigma(j)}}{\del \theta \del D_j} \right].
\end{align}

Therefore, we have 
\begin{align}
    \norm{\frac{\del \theta}{\del D_j}}_2^2 
    &= \sum_{k=1}^{\dparam} \left( [H^{-1}]_k\frac{\del \ell_{\sigma(j)}}{\del \theta \del D_j} \right)^2 \\
    &\le \sum_{k=1}^{\dparam} \norm{[H^{-1}]_k}^2 
    \norm{\frac{\del \ell_{\sigma(j)}}{\del \theta \del D_j}}^2 \\
    &= \norm{\frac{\del \ell_{\sigma(j)}}{\del \theta \del D_j}}^2 \norm{H^{-1}}_F^2 \\
    &\le \bound^2 \norm{H^{-1}}_F^2,
\end{align}
where the inequality is due to the Cauchy–Schwarz inequality. 

The remaining work is to bound $\norm{H^{-1}}_F$. Denote $\lambda_1, \ldots, \lambda_{\dparam}$ as the eigenvalues of $H$. Observe that 
\newpage
\begin{align}
    \norm{H^{-1}}_F 
    &= \sqrt{tr((H^{-1})^T H^{-1})} \\
    &= \sqrt{tr(H^{-1} H^{-1})} \\
    &= \sqrt{\sum_{i=1}^{\dparam} \frac{1}{\lambda_i^2} },
\end{align}
where the second equality holds due to $H$ being symmetric, hence $H^{-1}$ is also symmetric. Further, the last equality holds, because the eigenvalues for $H^{-1}$ are $\frac{1}{\lambda_1}, \ldots, \frac{1}{\lambda_{\dparam}}$. 
Since each $\ell$ is $\alpha$-strongly convex, we know that the minimum eigenvalue of the Hessian matrix $H$ is at least $n\alpha+n\lambda$. Therefore, we obtain 
$$
\norm{H^{-1}}_F \le \frac{\sqrt{\dparam}}{n \alpha + n\lambda}.
$$

Then, it follows that
\begin{align}
    \norm{\frac{\del \theta}{\del D_j}}_2 \le \bound \frac{\sqrt{\dparam}}{n (\alpha + \lambda)},
\end{align}

which implies 
\begin{align}
    \left|\frac{\del \theta_k}{\del D_j}\right| \le 
    \bound \frac{\sqrt{\dparam}}{n (\alpha + \lambda)},
\end{align}

and, finally, we have 
\begin{align}
    \left|\frac{\del \theta_k}{\del D}\right| \le 
    \bound \frac{\sqrt{\dparam (d+1)}}{\sqrt{n}(\alpha + \lambda)}.
\end{align}

\end{proof}

\subsection{Theorem 2}

\begin{customthm}{2}
If for all $z \in [0, 1]^{d}$, the loss function $\ell(\theta; z)$ is $\beta$-smooth in $\theta$, and $\norm{\frac{\del}{\del \theta \del z_i} \ell (\theta, z)} \le \bound$ for $i \in [d]$, 
then for $ \theta_k^{(t)}(D)$ defined by the $k$th entry of $ \theta^{(t)}$, which is iteratively computed by $\theta^{(t)} = \theta^{(t-1)} - \eta \g L(\theta^{(t-1)}; D)$ and $\theta^{(0)}=0$ with a learning rate $\eta > 0$, and $L(\theta; D) = \frac{1}{n} \sum_{i=1}^n \ell(\theta; (x_i, y_i)) + \frac{\lambda}{2} \norm{\theta}_2^2$, we have 
$$
\norm{\frac{\del \theta_k^{(t)}}{\del D}} 
\le \left( 1-(1-\eta \lambda -\eta \dparam \beta)^t \right) \frac{B_1 \sqrt{(d+1)} }{ \sqrt{n}(\lambda+\dparam\beta) }.
$$
\end{customthm}
\begin{proof}
Consider an iteration $\theta_k^{(t)} = \theta_k^{(t-1)} - \eta \g L(\theta^{(t-1)}; D)$. Taking a derivative with respect to $D_j$ for both sides, we have
\begin{align}
    \frac{\del \theta_k^{(t)}}{\del D_j} 
    &= \frac{\del \theta_k^{(t-1)}}{\del D_j} \\
    &-  \eta \left[ \lambda \frac{\del \theta_k^{(t-1)}}{\del D_j} 
    + \frac{1}{n} \sum_{i=1}^n \frac{\del^2 \ell_i}{\del \theta_k^2} \frac{\del \theta_k^{(t-1)}}{\del D_j} 
    + \frac{1}{n} \frac{\del \ell_{\sigma(j)}}{\del \theta_k \del D_j} \right] \\
    &= \left(1-\eta \lambda - \frac{\eta}{n} \sum_{i=1}^n \frac{\del^2 \ell_i}{\del \theta_k^2}\right) \frac{\del \theta_k^{(t-1)}}{\del D_j} - \frac{\eta}{n} \frac{\del \ell_{\sigma(j)}}{\del \theta_k \del D_j} \\
    &\le (1-\eta \lambda - \eta \dparam \beta) \frac{\del \theta_k^{(t-1)}}{\del D_j} - \frac{\eta}{n}B_1,
\end{align}
where the inequality follows due to $\left| \frac{\del \ell_{\sigma(j)}}{\del \theta_k \del D_j} \right| \le B_1$ and $\frac{\del^2 \ell_i}{\del \theta_k^2} \le \dparam \beta$. 

Observe that 
\begin{align}
    \frac{\del \theta_k^{(t)}}{\del D_j} + \frac{\eta B_1}{n \alpha}
    &\le (1-\alpha) \left( \frac{\del \theta_k^{(t-1)}}{\del D_j} + \frac{\eta B_1}{n \alpha} \right) \\
    &\le (1-\alpha)^t \frac{\eta B_1}{n \alpha}.
\end{align}
Therefore, we obtain
\begin{align}
    \left| \frac{\del \theta_k^{(t)}}{\del D_j} \right| 
    &\le \left( 1-(1-\eta \lambda -\eta \dparam \beta)^t \right) \frac{B_1}{ n(\lambda+\dparam\beta) },
\end{align}
which results in
\begin{align}
    \norm{\frac{\del \theta_k^{(t)}}{\del D}} 
    \le \left( 1-(1-\eta \lambda -\eta \dparam \beta)^t \right) \frac{B_1 \sqrt{(d+1)} }{ \sqrt{n}(\lambda+\dparam\beta) }.
\end{align}
\end{proof}

\subsection{Expressiveness of Deep ReLU Network}
\citet{yarotsky2017error} shows that continuous differentiable functions with a smaller upper bound of the gradient norm can be more efficiently approximated by deep ReLU networks. 
For completeness, we present the result here. 
\begin{theorem}
For any Lipschitz continuous function $f: [0, 1]^d \rightarrow \R$ with range $B$ and $\norm{\frac{\del f}{\del x}} \le L$ for all $x \in [0, 1]^d$, there is a ReLU network that is capable of expressing any such function within an arbitrary error $\eps$ and has no more than $c \ln( 2^{d+1} Bd(d+1) / \eps ) + 1$ layers and $d\left( 2^{d+1}dL / \eps + 1 \right)^d (c\ln( 2^{d+1} Bd(d+1) / \eps )+1)$ computational units. 
\end{theorem}
As we can see, if $L$ is smaller, the upper bound of the network size is smaller. The proof of this theorem is done by simply plugging in constants for Theorem 3.1 in \citep{yarotsky2017error}.

\section{Additional Experiments}

\subsection{SVM As the Base Model}
We summarize the results for dataset deletion and dataset addition experiments with Support Vector Machine (SVM) as base models in \ref{table:svmadd}.

\begin{table*}[h!]
\centering
\scalebox{0.8}{
\begin{tabular}{|c|c|cc|cc|cc|}
\toprule
\multirow{2}{*}{Dataset} &
  \multirow{2}{*}{Algorithm} &
  \multicolumn{2}{c|}{Parameter} &
  \multicolumn{4}{c|}{Utility} \\ \cmidrule{3-8} 
 &
   &
  $\norm{\widehat{\theta}-\theta^*}$ &
  Std &
  NRMSE &
  Std &
  Spearman Corr &
  Std \\ \midrule
\multirow{3}{*}{Iris} &
  \textbf{\AlgName} &
  \textbf{1.16E-01} &
  \textbf{1.53E-02} &
  \textbf{17.25\%} &
  \textbf{7.70\%} &
  \textbf{0.9214} &
  \textbf{0.0763} \\
 &
  Influence function &
  4.63E-01 &
  8.31E-03 &
  54.10\% &
  8.99\% &
  0.0143 &
  0.4427\\ \midrule
\multirow{3}{*}{SPAM} &
  \textbf{\AlgName} &
  \textbf{8.74E-01} &
  \textbf{1.58E-02} &
  \textbf{6.85\%} &
  \textbf{0.55\%} &
  \textbf{0.9826} &
  \textbf{0.0059} \\
 &
  Influence Function &
  1.47E+00 &
  7.78E-03 &
  48.96\% &
  3.23\% &
  0.1395 &
  0.2271 \\ \midrule
\multirow{3}{*}{HIGGS} &
  \textbf{\AlgName} &
  \textbf{5.38E-01} &
  \textbf{2.72E-02} &
  \textbf{7.52\%} &
  \textbf{1.16\%} &
  \textbf{0.9048} &
  \textbf{0.0442}\\
 &
  Influence Function &
  6.12E-01 &
  6.61E-03 &
  16.97\% &
  2.78\% &
  0.7274 &
  0.1377\\ \bottomrule
\end{tabular}
}
\caption{A summary of \AlgName results and baseline comparison results in the scenario of dataset deletion with SVM as the base model. The best results are highlighted in \textbf{bold}.}
\label{table:svmdel}
\end{table*}

\begin{table*}[h!]
\centering
\scalebox{0.8}{
\begin{tabular}{|c|c|cc|cc|cc|}
\toprule
\multirow{2}{*}{Dataset} &
  \multirow{2}{*}{Algorithm} &
  \multicolumn{2}{c|}{Parameter} &
  \multicolumn{4}{c|}{Utility} \\ \cmidrule{3-8} 
 &
   &
  $\norm{\widehat{\theta}-\theta^*}$ &
  Std &
  NRMSE &
  Std &
  Spearman Corr &
  Std\\ \midrule
\multirow{3}{*}{Iris} &
  \textbf{\AlgName} &
  1.68E-01 &
  1.15E-02 &
  \textbf{40.82\%} &
  \textbf{3.01\%} &
  \textbf{0.9292} &
  \textbf{0.0379} \\
 &
  Influence Function &
  \textbf{1.50E-01} &
  \textbf{6.72E-04} &
  63.29\% &
  3.88\% &
  0.1694 &
  0.2300 \\ \midrule
\multirow{3}{*}{SPAM} &
  \textbf{\AlgName} &
  \textbf{1.19E+00} &
  \textbf{1.77E-02} &
  \textbf{10.44\%} &
  \textbf{1.04\%} &
  \textbf{0.9848} &
  \textbf{0.0046} \\
 &
  Influence Function &
  1.62E+00 &
  3.27E-03 &
  55.03\% &
  2.01\% &
  0.8080 &
  0.0638 \\ \midrule
\multirow{3}{*}{HIGGS} &
  \textbf{\AlgName} &
  \textbf{7.23E-01} &
  \textbf{2.07E-02} &
  20.66\% &
  2.55\% &
  \textbf{0.8467} &
  \textbf{0.0378} \\
 &
  Influence Function &
  7.30E-01 &
  7.37E-03 &
  \textbf{17.34\%} &
  \textbf{2.41\%} &
  0.7095 &
  0.0828 \\ \bottomrule
\end{tabular}
}
\caption{A summary of \AlgName results and baseline comparison results in the scenario of dataset addition with SVM as the base model. The best results are highlighted in \textbf{bold}.}
\label{table:svmadd}
\end{table*}

The results in Table~\ref{table:svmdel} and \ref{table:svmadd} illustrate that the proposed \AlgName achieves similar performance with different base models (\ie LR and SVM) in terms of the accuracy of parameter prediction and strong correlation between the predicted utility and the actual utility of subsets.

\subsection{Larger Training Set}
\black{We increase the number of training data points to conduct larger-scale experiments to validate our proposed method.
The setting are summarized in Table~\ref{table:largesetting} and the results are summarized in Table~\ref{table:large}.}

\begin{table*}
\centering
\resizebox{\textwidth}{!}{%
\begin{tabular}{c|ccccc}
\toprule
Experiment & \shortstack{Total Training\\  Data Points} & \shortstack{Total Testing \\ Data Points} &  \shortstack{Size of \\ Staring Subset}  & \shortstack{Size of \\ Intermediate Subsets} & \shortstack{Size of \\ Ending Subset} \\ \midrule
Addition: MNIST-2000 & 2000 & 500 & 1000 & {[}1000, 1050, 1100, … {]} & 2000 \\
Deletion: MNIST-2000 & 2000 & 500 & 2000 & {[}1950, 1900, 1850, … {]} & 1000 \\
Deletion: MNIST-10000 & 10000 & 1000 & 5000 & {[}5100, 5200, 5300, …{]} & 10000 \\ 
Deletion: ADULT & 10000 & 1000 & 9250 & {[}9250, 9255, 9230, ...{]} & 10000 \\ \bottomrule
\end{tabular}
}
\caption{Setting for Experiments with Larger Training Set.}
\label{table:largesetting}
\end{table*}

\black{\textbf{MNIST-2000} \& \textbf{MNIST-10000}
We increase the number of training points to 2000 and 10000 for MNIST dataset and conduct the dataset addition and deletion experiments.
We construct 15000 training samples to construct the training set $\Phi$ for MNIST-2000 and 100,000 training samples for MNIST-10000.
Collecting 100,000 training samples takes roughly 3hrs.
We find \AlgName achieves similar effective results with very high Spearman correlation compared to the results of a training set with 300 data points.}


\black{\textbf{ADULT (\citep{adult2017})}
ADULT dataset (i.e., `Census Income' dataset) is a collection of  roughly 48000 records of personal income with 14 attributes (d=14).
The predefined prediction task is to determine whether a person makes over 50K a year as a binary classification problem.
We first sample 10000 data points with 9500 male records and 500 female records from the dataset as our full training dataset. 
Another 1000 points are selected for testing.
Instead of studying the dependency of the trained model on a randomly generated subset, we target utilizing \AlgName to investigate the effect of data collected from a subgroup (i.e. female). 
Therefore, we construct the 15000 training sample set $\Phi$ by removing a subset of these 500 points from the training set.
In the dataset deletion experiment, we also  remove subsets randomly generated by these 500 points from the training set.
The results in Table~\ref{table:large} show that \AlgName can accurately predict the trained model as well as the utility.
Therefore, it can be applied to study the effect of a subgroup within a large dataset on the training performance of a base model.}

\begin{table*}[h!]
\centering
\scalebox{0.8}{
\begin{tabular}{|c|c|cc|cc|cc|}
\toprule
\multirow{2}{*}{Dataset} &
  \multirow{2}{*}{Algorithm} &
  \multicolumn{2}{c|}{Parameter} &
  \multicolumn{4}{c|}{Utility} \\ \cmidrule{3-8} 
 &
   &
  $\norm{\widehat{\theta}-\theta^*}$ &
  Std &
  NRMSE &
  Std &
  Spearman Corr &
  Std \\ \midrule

\multirow{3}{*}{Addition: MNIST-2000} &
  \textbf{\AlgName} &
  1.44E+00 &
  5.00E-03 &
  \textbf{3.01\%} &
  \textbf{0.12\%} &
  \textbf{0.9993} &
  \textbf{0.0000} \\

 &
  Influence function &
  \textbf{1.41E-01} &
  5.00E-03 &
  60.10\% &
  0.31\% &
  0.9439 &
  0.0060\\ 
   &
  ParaLearn &
  1.10E-01 &
  \textbf{2.00E-03} &
  209.10\% &
  1.13\% &
  N/A &
  N/A\\ \midrule

\multirow{3}{*}{Deletion: MNIST-2000} &
  \textbf{\AlgName} &
  2.53E+00 &
  \textbf{2.38E-03} &
  \textbf{3.18\%} &
  \textbf{0.22\%} &
  \textbf{0.9912} &
  \textbf{0.0012} \\
 &
  Influence function &
  3.41E+00 &
  1.83E-01 &
  57.56\% &
  0.32\% &
  0.9439 &
  0.0060\\ 
   &
  ParaLearn &
  \textbf{1.87E+00} &
  4.57E-03 &
  125.32\% &
  0.79\% &
  N/A &
  N/A\\ 
     &
  Datamodel &
  N/A &
  N/A &
  311.20\% &
  7.37\% &
  -0.5854 &
  0.0271\\ 
  \midrule

\multirow{3}{*}{Deletion: MNIST-10000} &
  \textbf{\AlgName} &
  \textbf{2.30E+00} &
  \textbf{7.85E-03} &
  \textbf{1.19\%} &
  \textbf{0.08\%} &
  \textbf{0.9998} &
  \textbf{0.0002} \\
 &
  Influence function &
  3.08E+00 &
  1.18E-02 &
  56.57\% &
  0.55\% &
  -0.0302 &
  0.1309\\ 
   &
  ParaLearn &
  2.92E+01 &
  6.36E-03 &
  119.73\% &
  1.30\% &
  N/A &
  N/A\\ 
     &
  Datamodel &
  N/A &
  N/A &
  75.86\% &
  7.11\% &
  -0.0473 &
  0.1530\\ 
  \midrule

\multirow{3}{*}{Deletion: ADULT} &
  \textbf{\AlgName} &
  \textbf{3.27E-02} &
  \textbf{1.11E-03} &
  9.25\% &
  1.78\% &
  \textbf{0.9287} &
  0.0421 \\

 &
  Influence function &
  9.70E-01 &
  2.45E-02 &
  \textbf{8.86\%} &
  \textbf{1.68\%} &
  0.9146 &
  \textbf{0.0420}\\ 
   &
  ParaLearn &
  1.06E-01 &
  1.24E-03 &
  39.80\% &
  7.71\% &
  N/A &
  N/A\\ 
     &
  Datamodel &
  N/A &
  N/A &
  209.10\% &
  1.13\% &
  0.2454 &
  0.1030\\ 
  \midrule

\end{tabular}
}
\caption{\black{A summary of \AlgName results and baseline comparison results of experiments with larger training set. The best results are highlighted in \textbf{bold}.}}
\label{table:large}
\end{table*}

\subsection{Abaltion Study}
\begin{figure}[h!]
\centering
\includegraphics[width=0.6\columnwidth]{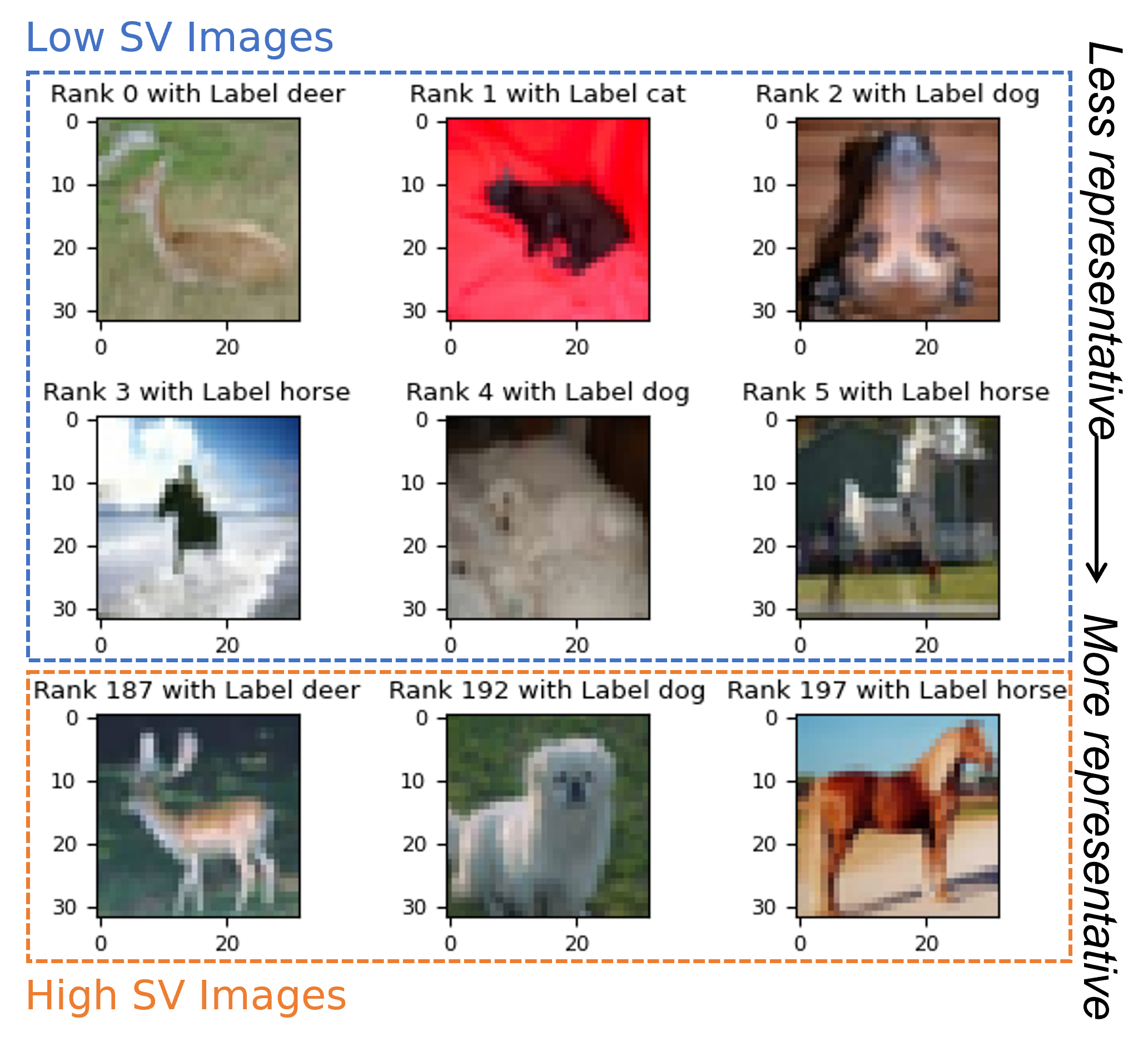}
\caption{Images with SV estimated by NN trained with transfer learning on a small subset of CIFAR-10.}
\label{fig:Abala_fix}
\end{figure}

\begin{figure}[h!]
\centering
\includegraphics[width=0.6\columnwidth]{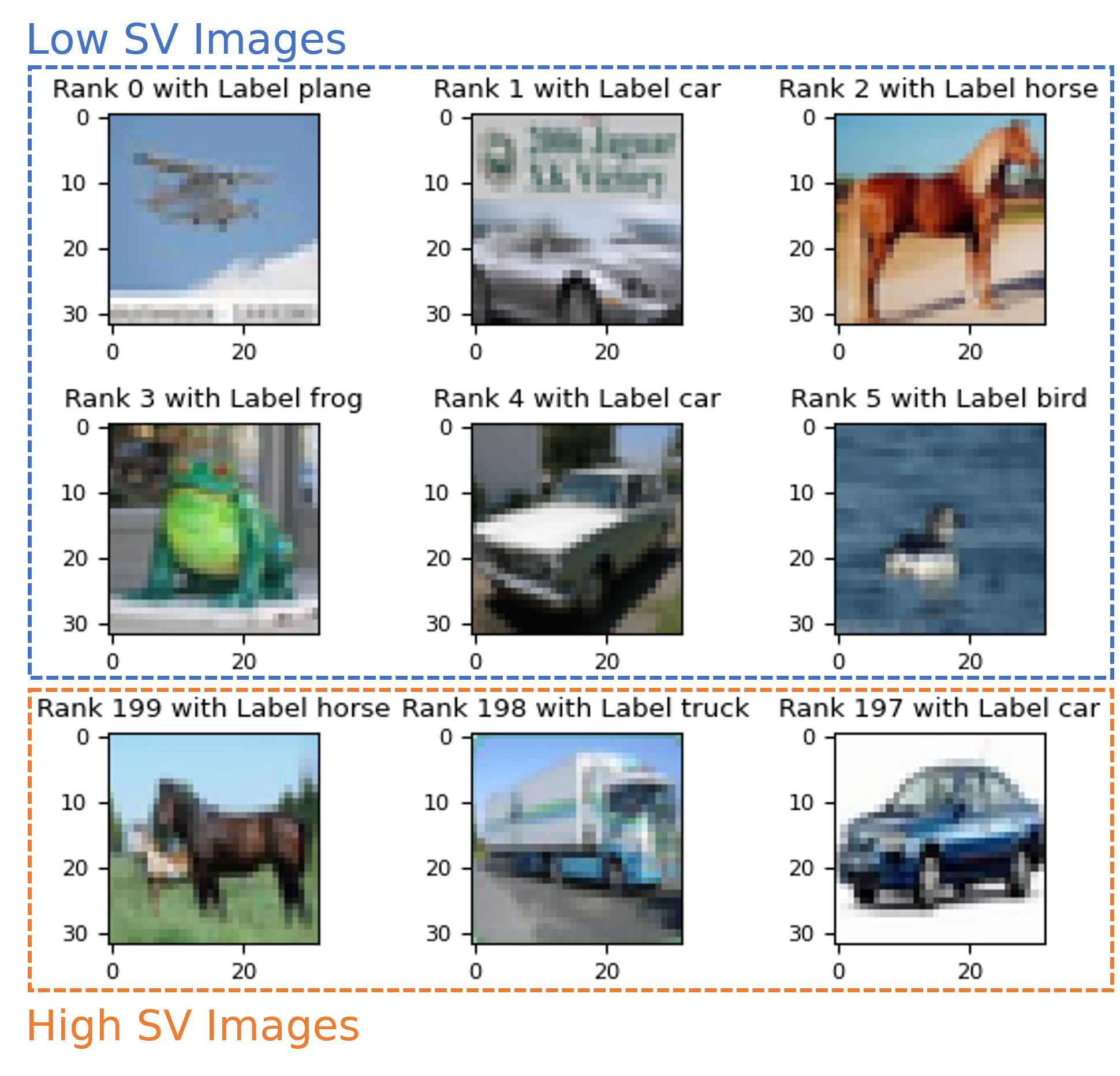}
\caption{Images with SV estimated by NN trained without transfer learning on a small subset of CIFAR-10.}
\label{fig:Abala_rdm}
\end{figure}

In the phase of offline training, we approximate the learning algorithm by fine-tuning the last layer of a pre-trained model.
We further perform an ablation study to validate the effectiveness of employing transfer learning in the large base model retraining.
A Resnet-18 model is pre-trained on CIFAR-10 dataset and the weight of the last layer is fine-tuned to a small subset (\ie with 200 samples) randomly selected from CIFAR-10.
To compare with fix embedding of feature extractors, we randomly initialize the weight of a Resnet-18 model and only adjust the weight of the last layer.
These two approaches are used to retrain the subsets of 200 samples in the permutation sampling to estimate the SV of these samples.
We select the samples with the lowest and highest SV estimated by these two approaches in Figure~\ref{fig:Abala_fix} and \ref{fig:Abala_rdm}.

As shown in Figure~\ref{fig:Abala_fix}, compared to samples in the same class (\ie deer, dog, and horse), those with lower SV (\ie in the blue box) are vaguer and less representative.
By comparing Figure~\ref{fig:Abala_fix} and \ref{fig:Abala_rdm}, we find that the same image (the third row and third column of Figure~\ref{fig:Abala_fix}) is estimated to obtain a high SV by transfer learning, but a low SV by random initialization.
However, this image is representative of the class of horse. 
Moreover, there is no significant difference in representativeness between the images with high and low SV estimated by NN with random initialization as shown in \ref{fig:Abala_rdm}, which might be caused by the low learning performance of NN during the subset retraining.
This indicates that retraining the large NN with transfer learning can effectively maintain the utility of training samples, which allows us to significantly reduce the model parameters to learn.
Therefore, \AlgName can be further utilized to learn the parameters of the large DNN models by transfer learning.

\end{document}